\documentclass[conference]{IEEEtran}
\usepackage[colorinlistoftodos]{todonotes}
\usepackage{amssymb}
\usepackage{longtable}
\usepackage{url}
\usepackage{textcomp}
\usepackage{lscape}
\usepackage{adjustbox}
\usepackage{array,ragged2e}
\newcolumntype{P}[1]{>{\RaggedRight\arraybackslash}p{#1}}
\graphicspath{ {./images/} }

\begin{document}

\title{Past, Present, and Future of Swarm Robotics}

\author{\IEEEauthorblockN{ Ahmad Reza Cheraghi}
	\IEEEauthorblockA{Technology of Social Networks \\
		Heinrich Heine University \\
		D{\"u}sseldorf, Germany \\
		ahmad.cheraghi@hhu.de \\ 
		https://tsn.hhu.de}
	\and
	\IEEEauthorblockN{ Sahdia Shahzad}
	\IEEEauthorblockA{Technology of Social Networks \\
		Heinrich Heine University \\
		D{\"u}sseldorf, Germany \\
		sahdia.shahzad@hhu.de \\ 
		https://tsn.hhu.de}
	\and
	\IEEEauthorblockN{Kalman Graffi}
	\IEEEauthorblockA{Honda Research Institute Europe GmbH \\
		Offenbach am Main, Germany\\
		kalman.graffi@honda-ri.de \\
		https://www.honda-ri.de
	}
}

\IEEEcompsoctitleabstractindextext{%
\begin{abstract}
Swarm Robotics is an emerging field of adapting the phenomenon of natural swarms to robotics. It is a study of robots that are aimed to mimic natural swarms, like ants and birds, to form a system that is scalable, flexible, and robust. These robots show self-organization, autonomy, cooperation, and coordination amongst themselves. The cost and design complexity factor is aimed to keep low, hence trying to form systems that are very much similar to natural swarms. The robots operate without any central entity to control them, and the communication amongst the robots can either be direct (robot-to-robot) or indirect (robot-to-environment). Swarm robotics has a wide range of application fields, from simple household tasks to military missions. This paper reviews the swarm robotics approach from its history to its future. It discusses the basic idea of swarm robotics, its important features, simulators, projects, real life applications and some future ideas. 
\vspace{-0.3cm}
\end{abstract}}

\maketitle             
\let\thefootnote\relax\footnotetext{\hspace{-0.3cm} }

\IEEEdisplaynotcompsoctitleabstractindextext

\IEEEpeerreviewmaketitle

\section{Introduction}
\begin{figure}[ht!]
	\centering
	\begin{minipage}[t]{0.45\linewidth}
		\centering
		\includegraphics[width=1.0\textwidth]{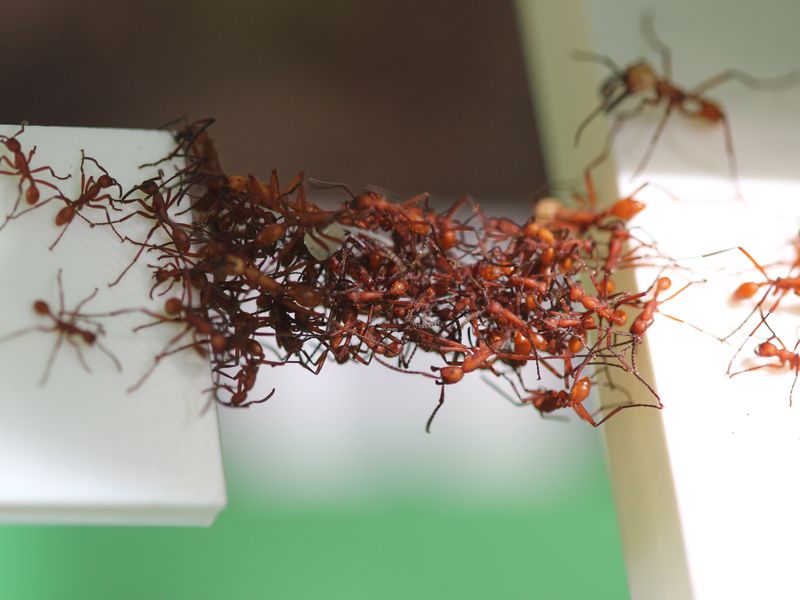}
		\caption{Ants building a bridge \cite{Antbridge}.}
		\label{ant}
	\end{minipage}
	\hfill
	\begin{minipage}[t]{0.45\linewidth}
		\centering
		\includegraphics[width=1.0\textwidth]{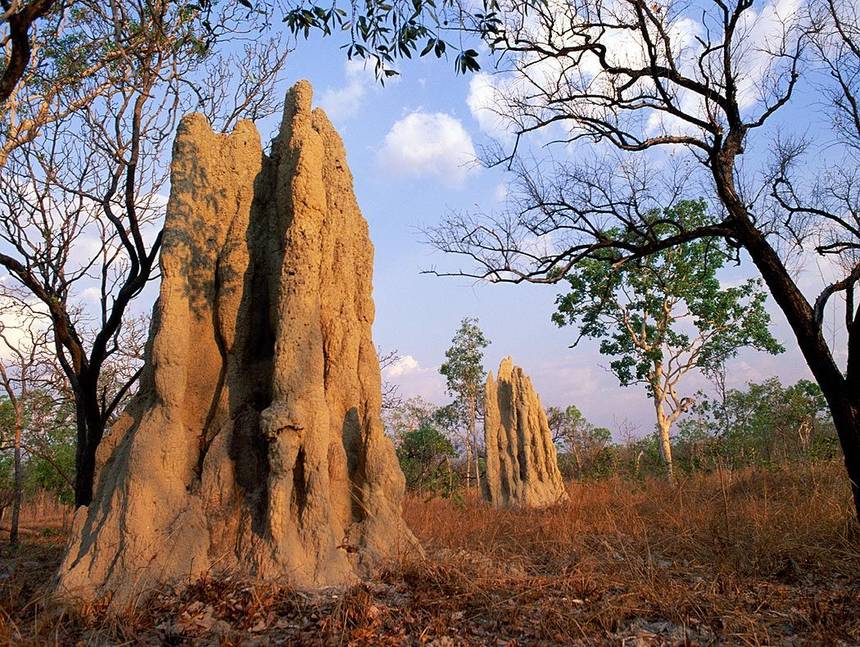}
		\caption{A mound built by termites \cite{Moundtermite}.}
		\label{mound}
	\end{minipage}
\end{figure}

The collective behavior shown by natural swarms like, honey bees, ants, fishes and others, has inspired humans to build such systems with robots, that can act in the most similar way as the natural swarms. These natural swarms can coordinate their simple behaviors and form complex behaviors with the help of which, they can accomplish tasks that are impossible for single individuals to perform. Swarm of ants can build bridges to cross large gaps, termites can build mounds that can be up to 30 feet high, fishes form shoals to protect them from predators and so on. Figure \ref{ant} shows a swarm of ants building a bridge to overcome a gap and figure \ref{mound} shows a mound built by termites. To realize natural swarm like systems in the field of robotics, it is first important to understand what a \textit{swarm} actually means. There are several definitions of a swarm in the literature. One simple and straightforward definition is given by \cite{barca2013swarm} :\textit{a large group of locally interacting individuals with common goals.} This means, it is aimed to build systems with swarms of robots that interact together and have some common goals to accomplish, just like natural swarms work together to accomplish common tasks. This paper deals with the idea of realising natural swarming into real life systems with robotic swarms.\\
This paper summarizes the research in the field of swarm robotics, from the starting till the perspectives of the future. The aim is to give a glimpse of the history of swarm robotics, the recent work in this field and the future plans. Section \ref{his} discusses the history of swarm robotics. It states, what the inspiration for this field was, what were the very first approaches and ideas and other historical aspects. Section \ref{srini} gives an overview of the swarm robotics field. It discusses the features, advantages, issues, tasks and application fields for swarm robotics. The present work, that includes different experimental and simulation platforms for swarm robotics, is described in section \ref{pres}. This section discusses the different types of simulators as well as real life applications in the field of swarm robotics. In section \ref{fut}, future perspectives, ideas and plans are discussed. Related surveys are mention in section \ref{relsur}. The last section concludes the work.  

\begin{figure}[ht!]
	\centering
	\begin{minipage}[t]{0.45\linewidth}
		\centering
		\includegraphics[width=1.0\textwidth]{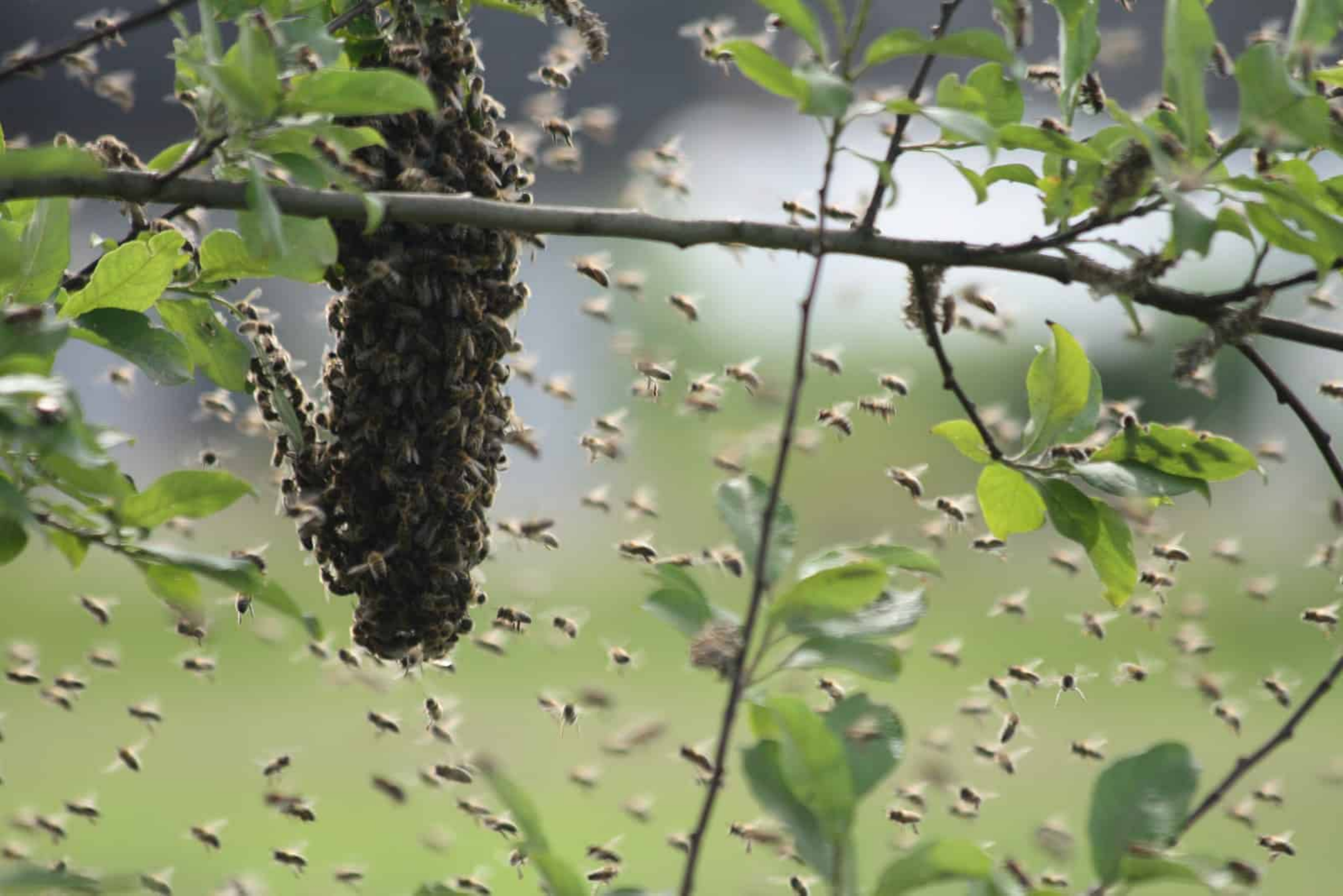}
		\caption{Swarming honey bees \cite{bswarm}}
		\label{bees}
	\end{minipage}
	\hfill
	\begin{minipage}[t]{0.45\linewidth}
		\centering
		\includegraphics[width=1.0\textwidth]{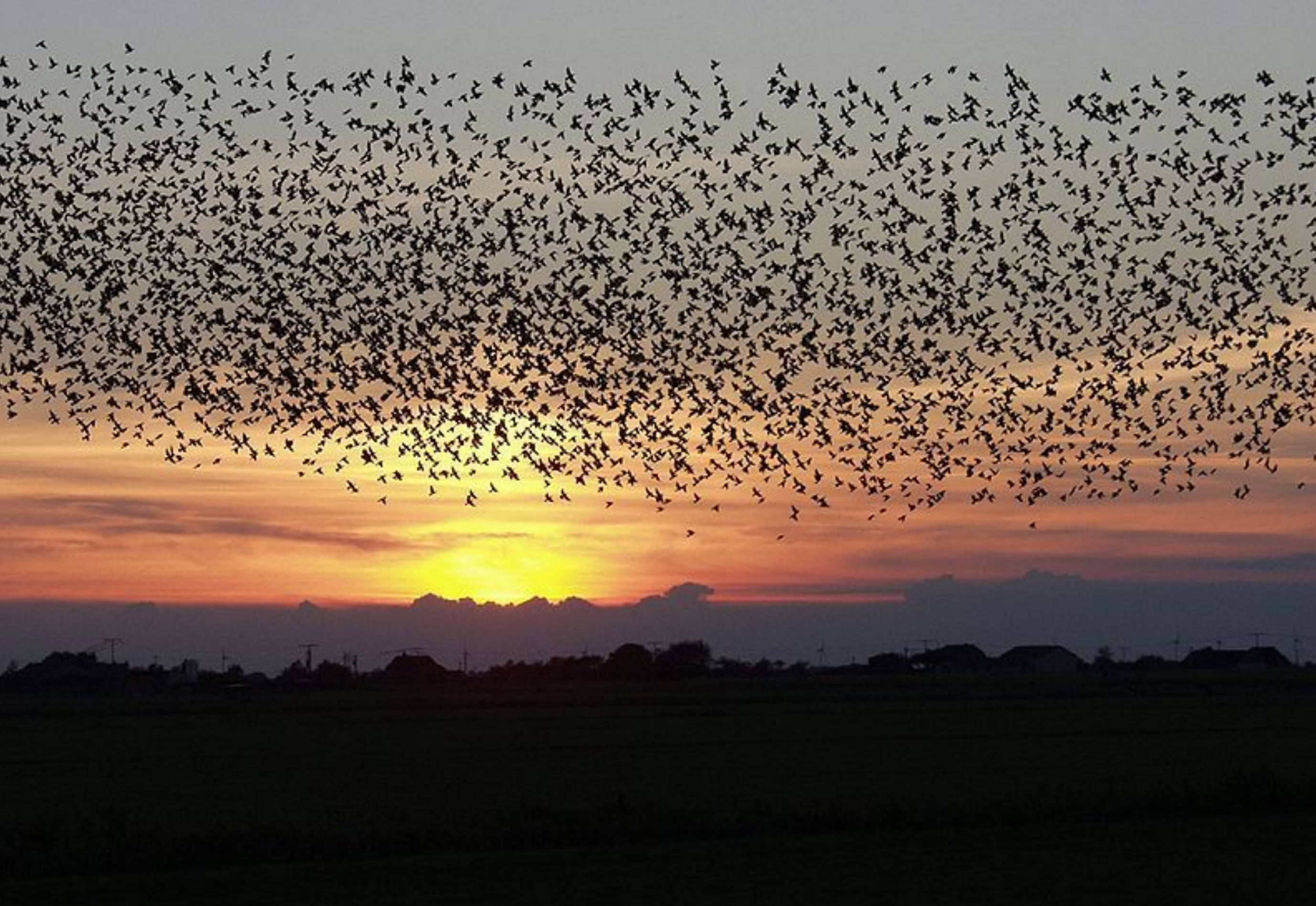}
		\caption{Birds flocking \cite{bgswarm}}
		\label{birds}
	\end{minipage}
	\hfill
	\begin{minipage}[t]{0.45\linewidth}
		\centering
		\includegraphics[width=1.0\textwidth]{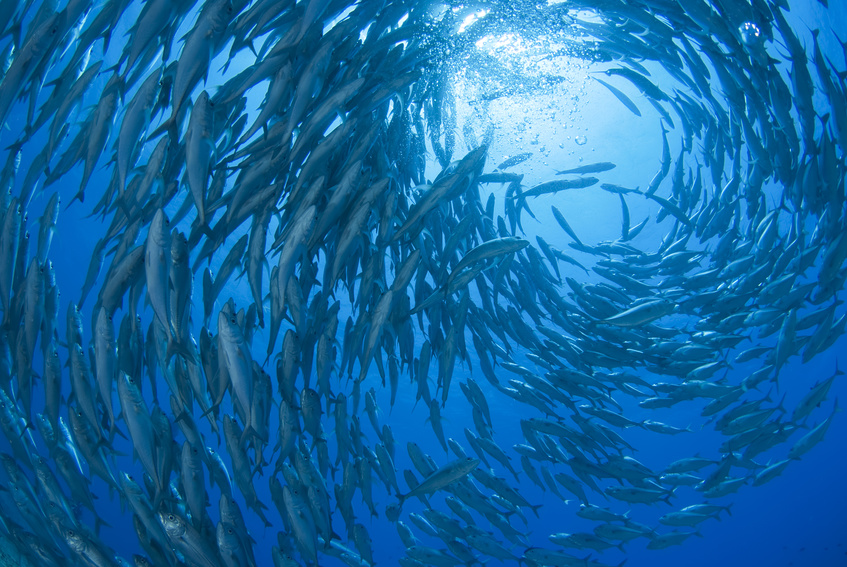}
		\caption{A fish shoal \cite{fswarm}}
		\label{fishes}
	\end{minipage}
	\hfill
	\begin{minipage}[t]{0.45\linewidth}
		\centering
		\includegraphics[width=1.0\textwidth]{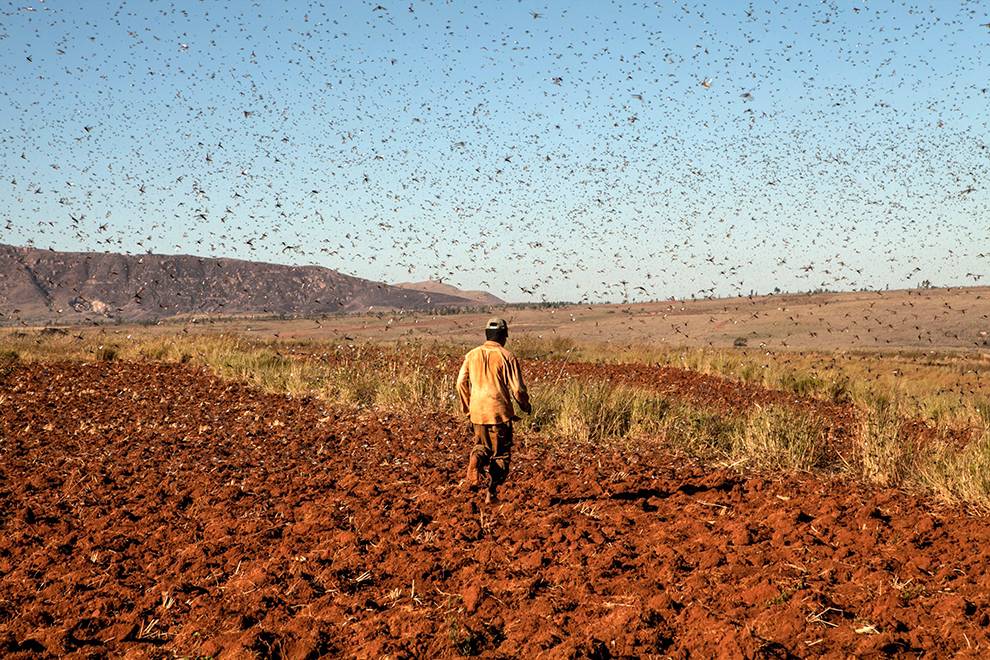}
		\caption{Locusts swarming \cite{lswarm}}
		\label{locusts}
	\end{minipage}
\end{figure}

\section{The History of Swarm Robotics}
\label{his}

The term 'swarm' in the context of \textit{robotics} is applied for the first time by G. Beni  \cite{65405beni} and Fukuda \cite{fukuda1988approach} in 1988. According to G. Beni, cellular robotics is a system composed of autonomous robots, that operate in a \textit{n}-dimensional cellular space, without any central entity. Additionally, they have limited communication among themselves, and they coordinate and cooperate to accomplish common goals. On the other hand, Fokuda uses swarm as a group of robots that can work together like the cells of a human body and as a result, they can accomplish complex goals. One year later G. Beni and J. Wang  \cite{beni1989swarm} introduces the term of \textit{swarm intelligence} in relation to cellar robotic systems . They claimed that cellular robotic systems are able to show 'intelligent' behavior via coordinating their actions. 

In 1993, C. Ronald Kube and Hong Zohng \cite{kube1993collective} constructed a multi-robot system that was inspired by the collective behaviours of natural swarms. At the same year, Gregory Dudek et al. \cite{583135dudek} define swarm robotics with respect to different features, including the size of a swarm, communication range amongst the robots in a swarm, communication topology, communication bandwidth, reorganisation rate of a swarm, abilities of swarm members and swarm homo- or heterogeneity. According to the authors 'swarm' is a synonym to multi-robotic systems, which is why it was still not clear what properties differ the term 'swarm robotics' from other robotic systems. 

In the early research on swarm robotic systems, the focus remained on the explorations of swarming behaviors in different species, like ants, birds, fish, and others. The researchers examined these behaviors and explored ways on how to realize these behaviors in robotic systems \cite{mataric1993designing, mataric1995issues, deneubourg1991dynamics, gage1992command, kube2000cooperative}. Additionally, research were driven by different inspirations, like the flocking of birds or colonies of ants. Natural swarms have always been the main motivation behind the idea of swarm robotics. Many studies and researches emulate different swarming behaviours like \textit{foraging, flocking, sorting, stigmergy}, or \textit{cooperation}. \cite{beckers1994local} and \cite{holland1999stigmergy} are two very old research works (1994 and 1999), that deal with the topic of \textit{stigmergy}. Stigmergy refers to the indirect communication amongst species and is introduced by \cite{grasse1959reconstruction} with reference to the behavior of termites. The first paper illustrates several experiments where mobile robots are responsible for collecting randomly distributed objects in an environment via stigmergy.  \cite{holland1999stigmergy} explores the feature of stigmergy and self organization amongst robots, having the same capabilities.

However, in 2004 G. Beni \cite{beni2004swarm} made another attempt to describe a swarm more precisely. According to him the robots in a swarm are simple, identical and self-organizing, and the system must be scalable, and only local communication is available amongst swarm members. These are the properties that are still considered as the basics of defining and distinguishing swarm robotic systems from other robotic systems. The robots used for the experimentation had a lot in common to social insects, for example the simplicity and the decentralization of the system. As a result, the word 'swarm' was used instead.  In the same time period, another research work \cite{csahin2004swarm} also dealt with the topic of swarm robotics. The author defined swarm robotics as \textit{" Swarm robotics is the study of how large number of relatively simple physically embodied agents can be designed such that a desired collective behavior emerges from the local interactions among agents and between the agents and the environment"}. He made some additions to the basic properties of a swarm robotic system. According to him, the robots must be autonomous, that means they should be able to interact with their environment and make decisions accordingly. Secondly, the swarms should consist of a small number of homogeneous groups, and each group should have a large number of robots in it.

Further, it is still not clear what size a swarm can or should be. G. Beni gives a brief definition to the size of a swarm as \textit{"It was not as large as to be dealt with statistical averages, not as small as to be dealt with as a few-body problem" }. According to the author the size of a swarm should be in the order of $10^2 - 10^{<<23}$. This means the number of members is greater than 100 and much less than $10^{23}$. The size also depends on the type of swarm robotic systems. If systems are for example, heterogeneous that involves robots with extra powers, then most probably the swarm size is not that big. This is due to the reason that the powerful robots can perform most parts of the task. On the other hand, a system that has very simple robots that cannot perform any significant task at their own, could have a huge size of swarm.

There have been several other definitions of swarms, all of them being similar in the way, that the main idea is to realize natural swarming, including their basic properties like, local interactions and coordination, into real life applications with swarms of robots. \cite{eberhart2001swarm} defined a swarm as:
\textit{A population of interacting individuals that optimizes a function or goal by collectively adapting to the local and/or global environment}

Researchers and developers aim to build robotic systems, known as \textit{swarm robotic systems}, consisting of large number of simple autonomous robots, that coordinate and cooperate, just like natural swarms, to attain simple to very complex tasks. The coordination and cooperation is gained via very simple rules. \textit{Swarm robotics} is the study of how to make the coordination and cooperation in large group of robots, possible. It deals with the realization of these swarming properties that are inspired from natural swarms, in the field of robotics. 

\section{Swarm Robotics: an Initial Approach}
\label{srini}
This section reviews the swarm robotic approach in details. Firstly, we present the different types of swarm i.e. swarm-behavior, -intelligence, and -engineering.
Next, some basic properties of swarm robotic systems. Like every other system, swarm robotics also comes with several drawbacks, these drawbacks along with the main advantages of swarm robotics are being enlightened in this section. The task areas and applications fields are also discussed. A comparison between swarm robotics and other robotic systems is given. Lastly, swarm robotics approach is classified along two axes, namely analysis and design. 

\subsection{Swarm-Types in Science}
As we know already the general meaning of a swarm is a group of species of the same kind that are moving. However, this term is used in science with different types, like \textit{biological swarms, swarm behaviors, swarm engineering} and \textit{swarm intelligence}. 

Biological swarms are the swarms found in nature, like swarms of ants, bird flocks, fish shoals, swarm of locusts etc.. These swarms work together in a coordinated way and form \textit{collective behaviors}. For example, cells and organisms that float or swim freely. \cite{kearns2010field} uses this swarm type for \textit{the bacteria becoming highly motile and migrating over the substrate}. Figure shows swarms in various species. 

The idea of thinking as a group so the whole swarm has the ability to decide or learn as one entity, is referred as \textit{swarm intelligence} \cite{beni1989swarm}. The motivation comes from natural swarms, like ants, honeybees or fishes, that work in groups for different purposes. They work together even if there is no direct communication amongst them. Ants look for food and left hints for others, namely the pheromones, to let others know about the food source. Similarly, fishes form fish shoals where a huge number of fishes migrate from one place to another. These fish shoals help them for example, to protect themselves from
predators. Thus, it is the intelligence of the whole swarm and not just a single entity from them.

The term 'swarm' is used for animals, insects and other species that show a \textit{swarm behavior} (see figure \ref{bees}, \ref{birds}, \ref{fishes} and \ref{locusts}). A swarm behavior is the combined intelligence of a swarm \cite{beni1989swarm}. It refers to the collective behavior that is demonstrated by fishes, birds, insects or other animals, in order to attain different goals. These individuals are mostly of the same species. However, mixed swarms of animals of different types and sizes may also be formed. Basically, we can conclude that a swarm is defined by the behavior of its beings, be it insects, animals or people.
 
The application of swarm based techniques is referred to \textit{swarm engineering}. The term "swarm engineering" was introduced by \cite{kazadi2000swarm}. He defined swarm engineering as \textit{a formal two-step process by that one creates a swarm of agents which complete a predefined task. The first step is to propose an expression of the problem which leads to a set of conditions on the individual agents which, when satisfied, will complete the task. The second step is to produce a behavior or set of behaviors for one or many robots which accomplishes these requirements.}  

\subsection{Properties}
\label{properties}
A swarm robotic system must exhibit several properties that are shown by natural swarms, to realize the idea of natural swarming in the most efficient way. G. Beni \cite{beni2004swarm} proposes the following properties: \\
\begin{itemize}
\item{\textbf{Flexibility}\\Swarm robotics aim to attain a verity of tasks. Here comes the feature of flexibility in focus. For the tasks, the system must be able to create various solutions by coordination and cooperation between robots. So, robots should find solutions by working together and be able to change their roles according to the given tasks. They should be capable of acting simultaneously according to the changes in their environment.}
\item{\textbf{Scalability}\\Scalability means that the systems must be able to work with different sizes of groups. There should not be a global number of robots present in a swarm, but the sizes may differ and accomplishing the task should still be possible and effective. The number of group members must not influence the performance of the system. So, swarm robotic systems should be able to operate with different number of members. The system should work effectively when the swarm size is small and it should support coordination and cooperation amongst the members, if the swarm size is large.}
\item{\textbf{Robustness}\\A system is referred as robust, if it has the ability to continue operating even if there are environmental disturbances or system faults. Environmental disturbances may include the changing of the surroundings, addition in the number of obstacles in the environment, weather changes and so on. Some of the system members can have a malfunction or can fail to perform. A swarm robotic system must be able to cope with such circumstances. In swarm robotic systems, individual robots are mainly very simple. This means that they cannot perform any significant tasks alone. So, if a system loses some robots it should not affect the overall performance of the system. The loss of individuals can be compensated by another member and the tasks must go on with the same level of efficiency.}
\end{itemize}
There are other features, that are not always exhibited by swarm robotic systems but are essential for differentiating them from other systems, like \textit{multi-agent-systems} and \textit{sensor-systems}. These properties are taken as a to-do list when designing swarm robotic systems. However, there are many systems that do not offer all of these features. These properties include: \\
\begin{itemize}
\item{\textbf{Autonomy}\\Acting randomly is known as autonomy. Robots that act and react on their own and make decisions by themselves are known as autonomous robots. They do not need a central entity to control their actions. }
\item{\textbf{Self organised}\\Robots coping with the environment and reorganizing themselves are called self-organizing robots. Self organisation is the most important aspect for swarm robotics. The main goal of most of the swarm robotic systems is to accomplish tasks in coordination and without a central entity.}
\item{\textbf{Self-assembly}\\self-assembly is the autonomous organization of robots into patterns or structures without human intervention or another central entity \cite{whitesides2002self}.}
\item{\textbf{Decentralized}\\Swarm robotics aims to achieve tasks without any central leader, due to many reasons like, it is difficult to control large swarms, a central control is a single point of failure, it is difficult to attain flexibility, scalability and robustness in centralized systems and many other reasons.}
\item{\textbf{Stigmergy}\\Stigmergy refers to the \textit{indirect communication} among robots. This form of communication is inspired by the \textit{pheromones}, that ants left on their way to the food sources, to signal other ants about some information, like the possible way to the food source.}
\end{itemize}
\subsection{Advantages and Issues}
Clearly there are a lot of advantages when tasks are accomplished by swarms rather than single entities. As we already discussed the motivation of swarm robotic systems comes from natural swarms, like ants, birds and fishes and they have different kinds of perspectives and goals for working in swarms rather than alone. Similarly, in swarm robotic systems there are various kinds of goals and tasks that need to be accomplished by grouping the robots. Some of many advantages of swarm robotic systems are mentioned below \cite{barca2013swarm}:
\begin{itemize}
\item{Robots are autonomous that can cope with their environmental changes.}
\item{Robots can combine their powers and abilities to form complex structures and offer unlimited features.}
\item{The systems are flexible. That means they can be applied in different fields and for a verity of tasks.}
\item{The systems are scalable, that means all robots can manage to obtain its goals no matter how big or small the swarm is.}
\item{\textit{Parallelism} makes the systems work more faster. Parallelism means tasks can be divided into sub tasks that can be allocated to different robots.}
\item{Robots are designed very simply, that means they are also cost effective.}
\end{itemize}
Like every other application and system, swarm robotic systems also have to deal with some issues. Regardless of all the features that swarm robotic systems offer like, scalability, robustness, stability, low cost and so on, the systems still deal with some drawbacks. The main aim for such a system is to realize them is real-world applications. Even if there are thousands of simulation and experimental platforms for swarm robotic systems, the realization into real world applications is not that straight and easy. Some of the issues that swarm robotic systems have to cope with are listed below \cite{mataric1995issues}:
\begin{itemize}
\item{The decentralized nature of swarm robotic systems make them a not so optimal choice for many applications.}
\item{Due to their \textit{autonomy} they will act to the changes in their surroundings individually and spontaneously. Even if the goal is to obtain tasks in a collective manner, the decentralization can result in single robots acting differently than the rest of the group.}
\item{The simple design and implementation of robots also makes it tough to design systems for real life applications in such a way that they achieve goals with a hundred percent guarantee.}
\item{For many real-life applications, global knowledge must be provided to robots.}
\end{itemize}
\subsection{Task areas and Tasks for Swarm Robotic Systems}
Applications of swarm robotics are widespread. They can be used for a verity of tasks where it can be very difficult or impossible for humans to achieve goals. There are many literature reviews \cite{khaldi2015overview, tan2013swarm, tan2013research} that indicate the areas of tasks for swarm robotics. Following are the main areas where tasks are and can be accomplished by swarm robotic systems \cite{csahin2004swarm}. \\
\begin{itemize}
\item{\textbf{Tasks in specific regions}\\
Swarm robotics is very beneficial to use in specified areas. Areas that are mostly large are filled with swarms of robots, and they act together to do work in this specific area. For example, collecting garbage from a city.}
\item{\textbf{Tasks in dangerous zones} \\
Attaining a task in dangerous regions is not practicable or better to say safe for human beings. It is useful to send robot swarms instead, in these dangerous regions. An example can be looking for hazardous objects in hazard fields or extinguish fire in a building. }
\item{\textbf{Tasks that can scale up and scale down} \\
Using swarm robotics in tasks that can scale up and scale down according to the circumstances is very beneficial because if a task is scaled up due to some reason, the number of members in a robot swarm can be increased and if a task is scaled down then the number of members can be decreased accordingly. For example, natural disasters can scale up very quickly.}
\item{\textbf{Tasks requiring redundancy} \\
Redundancy is a feature of swarm robotics and it is due to the factor of robustness that is shown by swarm robotic systems. This means that robots can cope with the failure of group members. They should continue with their work and the missing of some members should not have any influence on the performance.}
\end{itemize}
In the areas listed above, there are various types of tasks that can be accomplished by swarm robotic systems. These main tasks are mentioned below. 
\begin{itemize}
\item{\textbf{Forming shapes and patterns}\\
Robots can coordinate to form various shapes and patterns like stars, alphabets and so on. They deploy themselves in a specific manner and distance that results into the formation of patterns and shapes. SWARM-BOT \cite{dorigo2005swarm, inbookbot} is a project that deals with autonomous robot interactions and connection with each other to from different kinds of structures. A real-life application can be the forming of help messages that can be seen by rescue helicopters. Some of the studies on pattern formation include \cite{trianni2014evolutionary, maxim2009robotic}.}
\item{\textbf{Aggregation} 
Aggregation means clustering robots in a specific region. Under aggregation, it is aimed to group swarm of robots and let them attain tasks in an area. It helps robots to get close enough and interact to achieve goals. Farming by a swarm of robots is a possible real-life application by aggregating robots. \cite{dorigo2004evolving, soysal2007aggregation, soysal2005probabilistic} are some researches in aggregation tasks by swarm robotics. }
\item{\textbf{Coordinated movements} \\
In case of coordinated movements, the robots move in harmonized way. That means they move while preserving a specific form. A common example in natural swarms is the movement of fishes known as fish shoals. Through such movements many issues can be resolved, like avoiding obstacles or overcoming obstacles. Heavy objects can be transported easily and successfully by such a behavior. Some studies in this direction include \cite{ferrante2012self, hayes2002self}.}
\item{\textbf{Distribution of robots to cover an area}\\
The distribution of robots is the opposite of aggregating the robots. Here, the robots are scattered around in an environment in order to monitor their surroundings. This way the environment can be monitored in less time. \cite{6632825} introduces an algorithm for collective exploration. The idea is inspired by biological algorithms.}
\item{\textbf{Searching for specific sources}\\
The goal of searching items is one of the main tasks in swarm robotic systems. This idea is also inspired from natural swarms like ants that look for food collectively. The problem here is to find strategies that would result not only in finding items but also finding them in less time. \cite{article1} introduces a probabilistic model for robotic swarms in order to find food sources.}
\item{\textbf{Scanning an area and Navigation}\\
In order to get to specific targets, robots can scan their environment and navigate to target positions. A real-life example can be the searching of humans in a fire zone. Robots can scan the region of fire and navigate to their targets. Some research work includes \cite{nouyan2008path, ducatelle2011communication}}
\item{\textbf{Transporting objects}\\
Robots coordinate to transport objects if objects are not transportable by single robots. \cite{grobeta2008evolution} deals with the simulation of a swarm robotic system, where robots transport objects effectively by coordinating with each other.}
\end{itemize}
\subsection{Application Fields}
\begin{figure}[ht!]
	\centering
  \includegraphics[width=0.4\textwidth]{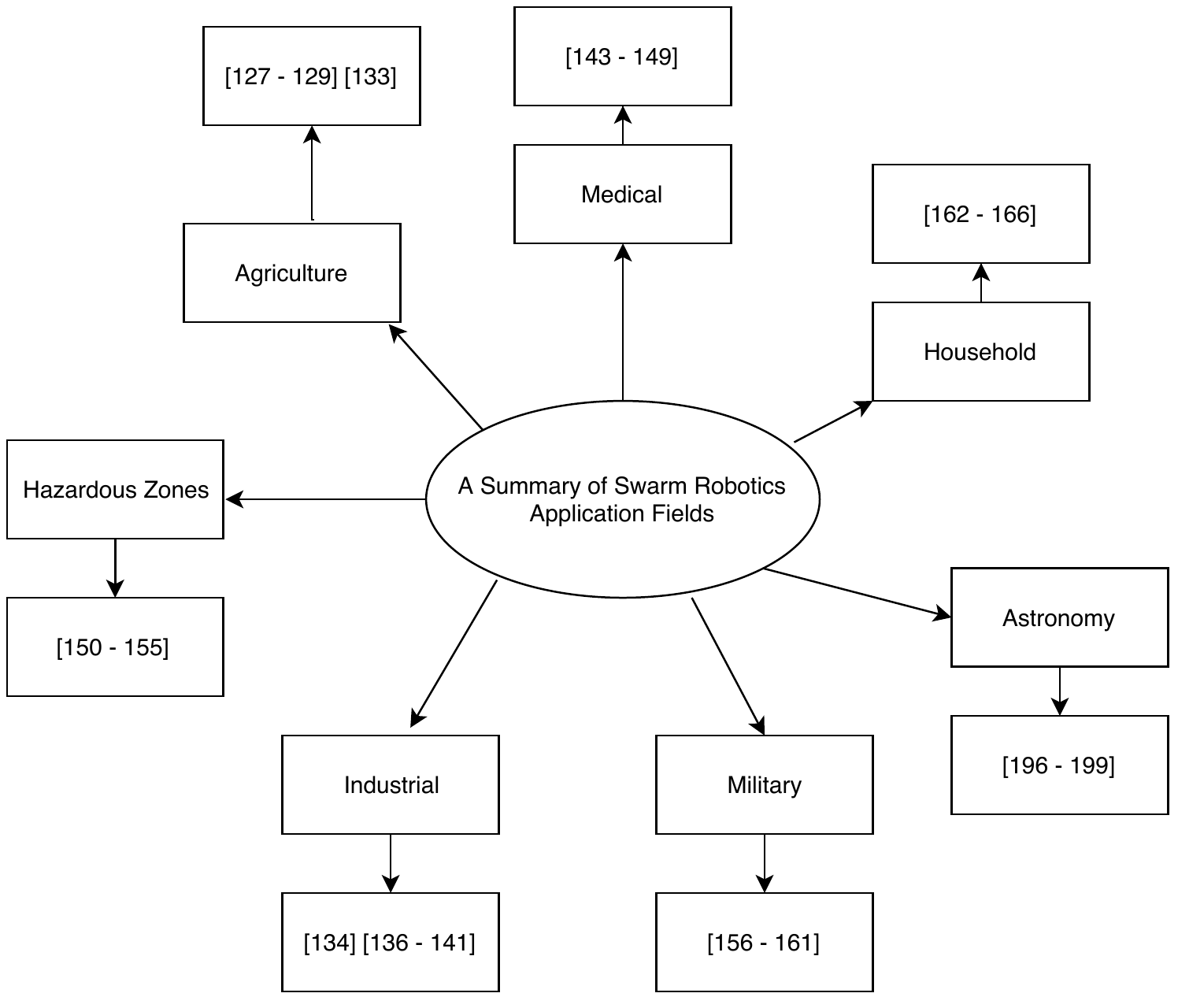}
	\caption{A summary of swarm robotics application fields}
	\label{applications}
\end{figure}

Swarm robotics is applied in many scenarios such as agriculture, medical, astronomy hazard zones and industrial areas. Some of the main application fields are listed below. In some fields, swarm robotic systems are already in use whereas in others the systems are still in research and underdevelopment phases. These application fields and some real-life swarm robotic systems in these areas are discussed in more details in section \ref{pres}. Figure \ref{applications} shows a summary of swarm robotics application fields. 
 \\
\begin{itemize}
\item{\textbf{Agriculture  }\\
In \textit{agriculture area}  swarm of robots are used to revolutionise farming and decreasing the work load of farmers [127 - 129] [133]. All farming tasks like harvest, sowing of seeds and so on can easily be done via robots.}
\item{\textbf{Industrial}\\
\textit{Industrial} fields also make the use of swarm robotics in various activities, like the dealing with chemicals [134] [136 - 141]. Here robots can be used instead of human beings in order to reduce any damage or harm to human workers.}
\item{\textbf{Military }\\
Swarm robotic systems can be of great use in the military [156 - 161]. For example, they can be used to detect and defuse bombs. This would exclude the need of human bomb diffusers. An army of robots can also be created to perform military tasks.}
\item{\textbf{Medical}\\
The use of swarm robots in the medical field is becoming very interesting and attractive over the time. Nanorobots can move into the veins and arteries to detect and cure various diseases like cancer cells. }
\item{\textbf{Astronomy}\\
Even though the research of swarm robotics in astronomy is not widely known, the use of swarm robotics in this field can be of great use [196 - 199]. For example, robots can be used to detect the effects of dark energy \cite{AStronomy}. }
\item{\textbf{Hazard zones}\\
Robots can be used to monitor dangerous areas to look for specific items like chemicals and toxins or survivors after a natural disaster [150 - 155]. Dangerous goods can also be transported via robots and mining work can be accomplished without any human interference. Such tasks are ideal to obtain with robots in place of humans so that there is no danger for human workers. }
\item{\textbf{Household}\\
Swarms of simple and small robots can be used for all day tasks, like cleaning [162 - 166]. There are many systems already in use. One such system is discussed later in section \ref{appreal}.}
\end{itemize}
\subsection{Swarm Robotic Systems and other Robotic Systems}
\begin{figure}[ht]
	\centering
  \includegraphics[width=0.4\textwidth]{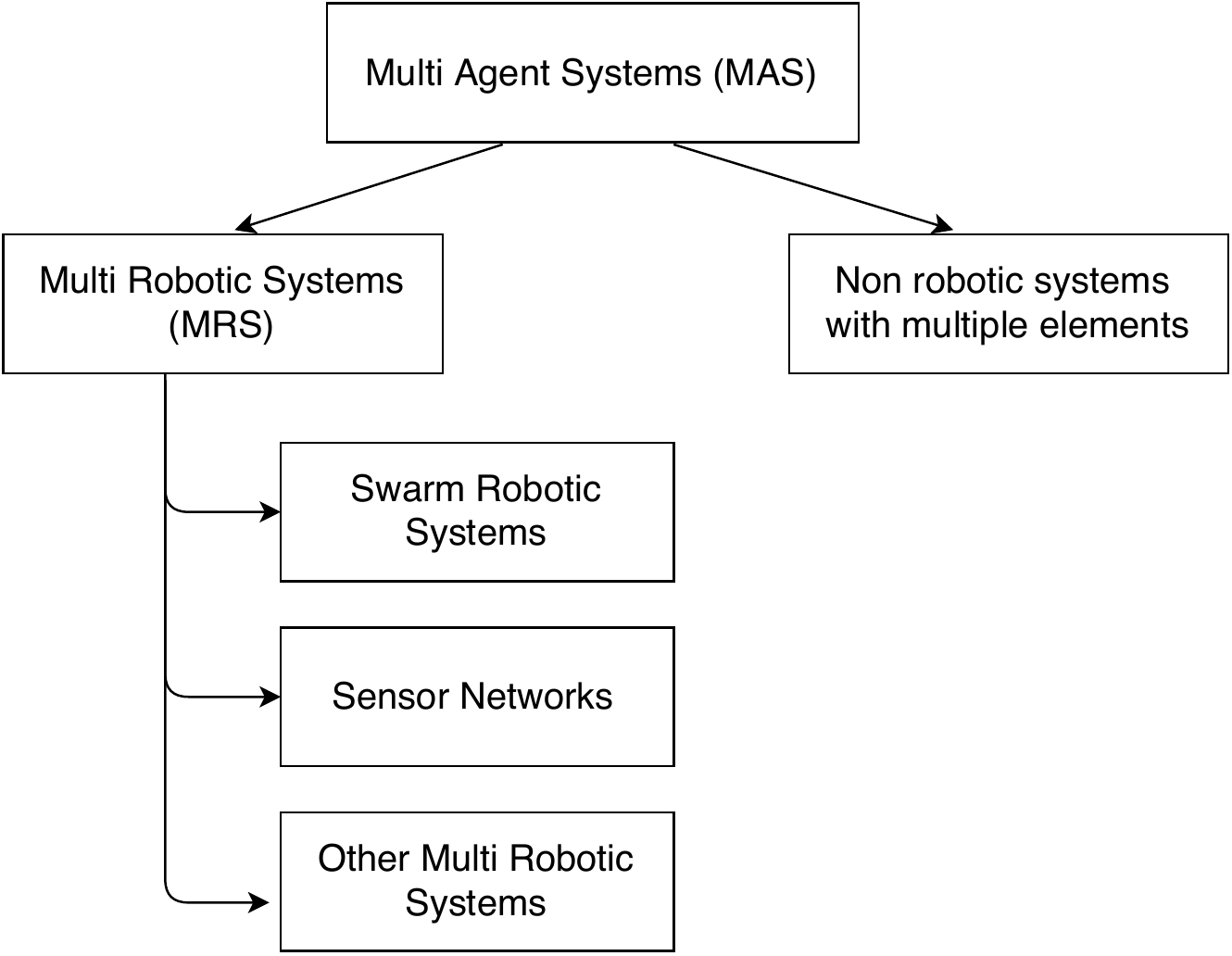}
	\caption{Robotic systems topology \cite{ronzhin2018interactive}}
	\label{robotclasses}
\end{figure}

There are many other systems besides swarm robotics that deal with multi-robots, be it real or virtual robots. But there are some differences between those systems and swarm robotic systems. Figure \ref{robotclasses} shows the robotic system topology. As you can see the main class is the \textit{multi-agent systems} (MAS) which is further divided into systems with robots and systems without robots. Under the robotic systems, that are the \textit{multi-robotic systems}, we have the \textit{swarm robotic systems}. To differentiate swarm robotics from other systems, a brief description of other similar systems and a comparison among swarm robotics and these systems is given in this section. Table \ref{robotsystems} shows the comparison and differences amongst the systems.
A \textit{multi-agent system (MAS)} consists of multiple software \textit{agents} that are autonomous, flexible and interact in an environment to perform tasks. In the literature we can find various definitions for an agent. \cite{maes1995artificial} defines an agent as \textit{computational systems that inhabit some complex, dynamic environment, sense and act autonomously in this environment, and by doing so realize a set of goals or tasks for which they are designed}. The sharing of information between these agents is over a network. In other words, it refers to a system whose hardware or software components are distributed on networked computers, and they communicate and coordinate their actions through message exchange \cite{coulouris2005distributed}. The information sharing over a network is the point that differentiates multi-agent systems from other robotic systems. The agents either work together to achieve a goal or they can also work for different goals. One common feature in MAS is that the agents can communicate with each other. \\
\textit{Multi-robotic systems (MRS)} represent a collection of robots that aim to perform a common task. Using multi-robots to perform tasks is a solution when single robots are not able to accomplish these tasks \cite{shi2012survey}. So, the use of multiple robots is required if a robot alone is unable to perform a specific task. For example, if a heavy item needs to be transported, it can be a difficult or impossible task for a single robot. Due to some specific activities, it is a better or an ideal solution to use a group of simple designed robots rather than complex single robots. What differentiates MRS from swarm robotics, can be seen in table \ref{robotsystems}. Most multi-robotic systems are heterogeneous. This means that robots have different capabilities. The size of population in MRS is small as compared to swarm robotic systems and the control of this population is usually under one entity, that makes the system centralized. The robot that acts as a leader has more capabilities than others. This can increase the costs and hence making the system not so cost efficient. Like in swarm robotics, the robots in MRS are also mobile but in contrast to swarm robotics, they are mostly familiar with their environments. When it comes to flexibly, scalability and robustness, swarm robotic systems are much more flexible, scalable and robust than MRS. This is due to the reason that swarm robotic systems have large populations that decreases the chances of bad performance in case of robot failures. Robots are able to cooperate and coordinated even in large swarms and such systems can be applied for different kinds of tasks.\\
According to \cite{shakkottai2003cross}, a \textit{distributed sensor network} consists of hundreds to several thousands of sensor nodes distributed over a field of application. The sensor nodes monitor their environment with the help of sensors and the communication between these members is wireless. Each node can actively participate in the monitoring and also communicate with other members. So, there are no restrictions for the nodes to take part in only one activity. Large amount of sensors are placed into an environment, and they coordinated with each other to generate information about their environment. This information can be for example the atmospheric density in that area. The medium used to communicate is wireless, for example, radio waves, infrared waves or optical media \cite{akyildiz2002survey}. The collected data from each sensor is send to a base station, known as a \textit{sink}. For the management of the data, a \textit{task manager} is responsible and the communication between the sink and the task manager is via Internet or satellite. 

\begin{table*}[tb]
 \center
\begin{tabular}{|l|p{2.6 cm}|p{2.7 cm}|p{2.6 cm}|p{2.6 cm}|}
\hline\textbf{Robotic Systems}& \textbf{Swarm Robotic Systems} &
 \textbf{Multi-Robotic systems}& \textbf{Multi-Agent Systems}& \textbf{Sensor Networks} \\
\hline \hline
Number of members & Large (as compared to other robotic systems) & Small (as compared to swarm robotic systems) & Small (as compared to swarm robotic systems) & Large (as compared to MAS and MRS)\\
\hline 
Design and implementation of robots & Very simple. Single robots are unable to do anything significant & Single robots can perform significant parts of a task & Single robots are able to perform significant parts of a task & Nodes can be designed simple or complex \\
\hline
 Self-organisation & Yes & Yes & Yes & Yes\\ 
\hline
System Control (centralized or decentralized) & decentralised & Both & Both & Both \\ 
\hline
Homogeneity or heterogeneity & Mostly homogeneous & Mostly heterogeneous & Both & Homogeneous \\
\hline
Autonomy & Yes & No & No & Yes\\
\hline
Environment & unstructured (unknown) & structured and unstructured (known and unknown) & structured (known) & structured (known)\\ 
\hline
Movement & Yes & Yes & Mostly not & No\\
\hline
Robustness & yes (high) & Yes & Yes & Yes \\ 
\hline
Scalability & yes (high) & Yes (low) & Yes & Yes \\ 
\hline
Flexibility & yes (high) & Yes (low) & Yes & Yes \\
\hline
Cost & Low & Medium & Medium & High\\
\hline
\end{tabular}
 \vspace{0.3cm}
\caption{Differences and similarities between swarm robotic systems and other robotic systems.}
\vspace{-0.7cm}
\label{robotsystems}
\end{table*}

\subsection{Swarm Robotics Classification}
Swarm robotics has a large rang of fields, where it can be applied. The tasks done by swarms of robots vary from mini tasks to macro tasks. That means, the task may be of just transporting one particle from one place to another or transporting huge building blocks from one place to another. All tasks performed can be categorized under different types of \textit{swarm robotic behaviors}, such as swarms making decisions according to the circumstances or the making of decisions is preprogrammed, does the task need collective transport or collective exploration? And so on. All these possible classes of collective behaviors are discussed in this section. These behaviors and groups also need efficient modeling and analysis. In order to do so, two types of methods, namely the \textit{Design methods} and the \textit{Analysis methods} are presented \cite{brambilla2013swarm}. \\
Analysis refers to the detailed systematic study and examination of components, construction, and structure of a system. To use a system in practical life, it is indeed a very important step to analyze them. It helps in making decisions about the system. For example, what component needs to be advanced, removed, or may be added and what features do we need additionally and so on. The reactions of robots are often spontaneous that is why it is difficult to illustrate what the next step is going to be. For example, if there is an environment filled with obstacles, the robots may choose to follow another path, or they can also jump on the obstacles if they are designed accordingly and have such types of features. Even if the overall behavior and basic features are known, but still there can be parameter values that are unknown. The classification model presented below is inspired by \cite{bayindir2007review, csahin2008swarm, tan2013research} and \cite{brambilla2013swarm}.
\subsubsection*{Methods for Analysis} 

\paragraph{Microscopic Approaches}
In microscopic approaches the modelling of robots is done on microscopic level, that means on individual level. The characteristics features and behaviors of every single robot and the environment is modelled analytically. This analyzing also includes the interactions among the robots and the interactions between robots and their environments.  

\paragraph{Macroscopic Approaches}
Macroscopic approaches models on behaviors on swarm-level instead of modelling on individual robot level. So, the whole system is not analyzed by first analyzing the robots individually, but the whole system is analyzed only once in order to obtain the system state. One can see that such an approach is faster than the microscopic approach, but through such modelling, only a rough global view of the system can be obtained, and small experimental failures cannot be captured. 

\paragraph{Sensor based Approaches}
Just like microscopic modelling, sensor-based modelling is also on individual level. As the name says, sensors are used as the main component for the modelling. Along with the sensing, the actuation of individual robots, interaction between robots and interactions between robot and their environments is also modeled. The aim is to make these interactions simple and realistic \cite{bayindir2007review}. These features make this approach different from other approaches. Interactions must be simple because their complexity is of high importance when experiments are conducted to test the scalability of the system. Being realistic is also an advantageous feature for swarm robotic systems. 

\paragraph{Modelling via Swarm Intelligence Algorithms}
Swarm intelligence algorithms can be used for many purposes, like robot controlling, robot research, swarm behavior predictions and modelling of swarm robotics. \cite{inbook} gives a survey of swarm intelligence algorithms, that are inspired by ants, bees, fireflies, glow-worms, bats, monkeys, lions, and wolves. The \textit{particle swarm optimization} (PSO) algorithm is often used for modelling and analysis purposes \cite{eberhart1996computational, eberhart1995new, kennedy1997particle}. 

\paragraph{Real-Robots Analysis Approaches}
Instead of simulating all real-life robotic situations, real robots can also be used for modelling and analysis purposes. One can test collective swarm behaviors on real robots, which seems to be a difficult task when there are a huge number of robots and behaviors, but we can accomplish good and reasonable results. The difficulty is mostly in the case of speed. Using models and simulators are usually faster and easier than real robots. This is also the reason that there are a lot less research for real robot analysis than models and simulator analysis. 

\subsubsection*{Methods for Design}\mbox{} \\
\textit{a) Behavior based Approaches} \\
Behavior based approaches, also known as \textit{ad-hoc} approaches \cite{csahin2008swarm} are the most common way of designing swarm robotics systems. In this type of approach, the desired behavior of each individual robot is achieved in an iterative way, by first implementing an inspired behavior manually and then enhancing it, whenever needed. \cite{brambilla2013swarm} has divided the behavior based approaches into further two categories, namely the \textit{probabilistic finite states machines design} and \textit{virtual physics-based design}. The use of \textit{finite state machine} (FSM) in swarm robotics is very common due to its capability of changing from one state to another. This feature helps in achieving sensor or input-based behaviors of swarm robots. Input base behavior means that the robots act according to the given inputs, for example, obstacles on the way. In probabilistic finite machine (PFSM), the state changes have transition probabilities. These probabilities can change over time or stay the same, as needed. Many swarm behaviors are generated using probabilistic FSM. These behaviors include, \textit{foraging} \cite{liu2007modelling}, \textit{task allocation} \cite{liu2007towards, labella2006division}, \textit{aggregation} \cite{soysal2005probabilistic} and \textit{chain formation} \cite{nouyan2008path}.

\textit{b) Virtual physics-based Design} is a virtual design, where every robot is a virtual particle that performs virtual actions. These virtual actions are reactions to the virtual forces, motivated from natural physics laws \cite{spears2004distributed}. It is assumed that the robots have some specific features, like sensing of their virtual environment. They can sense other robots in their environment, identify individual robots and also calculate their relative distances and positions. This type of approach is used to design swarm behaviors, like pattern formation and coordinated motion. 

\textit{c) Automatic Approaches}\\
In automatic approaches, swarm behavior is not implemented manually by the developers. Swarm of robots can learn different types of behaviors, via various scenarios in their environment, like other robots coming in way, obstacles on the way, and so on. These design methods are a part of \textit{machine learning}, \textit{deep learning} and \textit{reinforcement learning}. 

\section{The Present State}
\label{pres}
Swarm robotics is one of the very popular topics of research nowadays. There is a lot of research work on this topic. The projects take their inspiration from natural swarms and try to build a platform that works in the most similar way as natural swarms. The main properties of these systems are that they have mostly very simply programmed robots, the populations are of large sizes and the costs are low to medium. The costing factor is kept low because single robots are very simply designed and have very less capabilities. So, the tasks are accomplished by the coordination and autonomy of the robots. Three projects are elaborated in \ref{proj}. Table \ref{projectscomp} gives an overview of several swarm robotic projects. \\
To deploy a system in real-life applications, it first needs a lot of testing to make sure everything is working perfectly, and the aimed tasks are accomplished efficiently. But the testing of these systems with real robots can be very expensive. This is the main reason why simulation platforms are an ideal way to analyze and model swarm robotic systems. Swarm robotic systems can be modelled with simulators and changes can be made easily, in less time and money. Various kinds of experiments can be conducted, and the results can be analyzed. When the desired results are acquired, they can then be transferred to systems with real robots. Some simulation platforms are discussed in section \ref{simu}. A comparison of several simulators is given in table \ref{similatorcom}.\\
Swarm robotic systems has a vast range of applications \cite{Camazine:2001:SBS:601161, navarro2013survey, schwager2009theory, hayes2001swarm, brutschy2014self}. There are various simulation platforms and real robot systems established to research on and test the significance of swarm robotics in real-life scenarios. Some real-life application platforms are discussed in section \ref{appreal}. 

\subsection{Projects}
\label{proj}
\subsubsection*{SWARMBOT}
SWARM-BOT \cite{dorigo2005swarm, inbookbot, dorigo2004swarm} is a project on swarm robotics sponsored by the Future and Emerging Technologies program of the European Commission (IST-2000-31010). This project aims to study the design and implementation of self-organizing and self-assembling swarms of robots. Self-assembly is a feature where robots can make physical connections and form patterns and shapes together to cope with difficult situations and overcome obstacles. The system consists of small autonomous robots known as \textit{s-bots}. Multiple sensors, such as light or infrared sensors, traction sensors and actuators are equipped in the \textit{s-bots}. They have limited computational capabilities, and to communicate with other members, basic communication devices are also available to these robots. In addition to these basic elements, they are also equipped with physical connection mechanisms, that helps them to attach as well as detach from other members. \textit{S-bots} connect and form \textit{swarm-bots} that have higher capabilities as compared to a single \textit{s-bot}. So, a \textit{swarm-bot} be a single entity with many parts that can be detached at any time according to environmental circumstances. Even a single \textit{s-bot} can be used to accomplish simple tasks according to its capabilities. These simple tasks may be to grasp small items, navigate autonomously or transport small objects. If a tough task comes forward, that it not possible to obtain, it attaches itself to others and forms a \textit{swarm-bot}. The configured \textit{swarm-bot} can change its shape and connect or disconnect from \textit{s-bots}. An additional component is the \textit{s-toy}, that can be used as an object to transfer or as a symbol to locate targeted items. The physical appearance of a \textit{s-toy} is the same as of a \textit{s-bot}. It can attach and detach from \textit{s-bot}s. A \textit{s-bot} is made up of two parts, an upper tracking part and a lower part which consists of wheels. The tracking part can rotate with respect to the upper body part of the robot, that is known as the \textit{turret}. The rotation is done via motorized link of the turret. In addition, two grippers namely, the rigid gripper and flexible gripper are mounted on the turret. Through these grippers \textit{s-bot}s can join with other \textit{s-bot}s.

\subsubsection*{Marsbee}
Marsbee \cite{bluman2017marsbee} is a NASA funded project which aims to send a swarm of robotic bees on Mars. The goal of these marsbees is to explore Mars, that is, for humans very difficult to do themselves. The exploration of the Red planet began a long time ago, but the area already investigated is not a large section. Hence, NASA decided to send robotic bees to explore the planet. These robotic bees known as, marsbees, are a size of a bumblebee with flapping wings attached to them. The size of their wings is larger than of bumblebees. To explore the area efficiently, sensors are integrated in each marsbee. Through these sensors, they can investigate their environment, for example, measuring the temperature or humidity, look for types of obstacles, explore the type of surface, search for water and food resources and so on. A mobile base station is used to launch and recharge the marsbees. It is also used as a communication interface between marsbees and the main base. Marsbees communicate wireless between themselves and with the base station.  \\
The project is divided into two phases. In the first phase the design, movement and weight of the flapping wings will be determined. These aspects are necessary in order to make the marsbees able to fly easily and with enough power, on Mars and cope with the environmental changes. The second phase addresses the mobility, remote sensing abilities and optimisation tools. 

\subsubsection*{Kobot}
Kobot \cite{turgut2007kobot} is a circular shaped, cheap, small and expendable robot. These properties make it very suitable to swarm robotics platform. It has size of a CD, a diameter of 120 mm and a weight of 350 grams. It is equipped with batteries, motors and sensors. The body of a kobot consists of two parts, namely a \textit{cylindrical base} and a \textit{cylindrical cap}. The batteries, motors and sensors are placed in the cylindrical base. The cylindrical cap covers the cylindrical base. Kobot consists of two high quality DC gear-head motors, that are used for the locomotion, and a battery with up to 10 hours battery life. A modulated infra red system (IR system) is also equipped with the help of which obstacles and robots can be distinguished. The system uses a wireless communication platform, namely the  
IEEE802.15.4/ZigBee protocol. Via this protocol, parallel programming of robots and peer-to-peer communication is made possible. 

\begin{table*}[tb]
 \center
\begin{tabular}{|l|P{6.3 cm}|P{6.3 cm}|P{2.0 cm}|}
\hline\textbf{Projects}& \textbf{Objectives} & \textbf{Basic Properties} & \textbf{Researches}\\
\hline
\hline
SWARM-BOT & To form simple, reliable, flexible, scalable, self organized and self assembling micro-robotic systems & Robots show robustness, flexibility and are able to solve complex problems via self organisation &\cite{sahin2002swarm, mondada2003swarm, dorigo2004evolving, mondada2004swarm, dorigo2004swarm}\\
\hline
Swarmanoid & The main goal is to design a heterogeneous distributed swarm robotic system that operates in 3-dimensional human environments & In addition to s-bots, Swarmanoid consists of hand- and eye-bots that can climb objects and fly, respectively & \cite{dorigo2009swarm, pinciroli2007swarmanoid, ducatelle2011self, dorigo2013swarmanoid, decugniere2008cart} \\
\hline
I-SWARM & The goal of the I-SWARM project is to build the largest robotic swarm that consists of up to 1000 mini robots of size 3*3*3 mm and a single robot looks and moves like an insect. But it consists of various modules that enables it to perform significant tasks. These modules include, power, electronics, locomotion and communication module & \cite{woern2006swarm, seyfried2004swarm}\\
\hline
SensorFly & To build an aerial mobile sensor network of robots that can perform monitoring in indoor emergency situations & The aerial robots are low-cost, autonomous, and are capable of 3D sensing, obstacle detection, path identification and adapting to network disruptions & \cite{purohit2012demo, purohit2011sensorfly}\\
\hline
Marsbee & Exploring Mars & Consists of a colony of small flying robotic bees that can sense their environments via sensors. There is a charge station  where the marsbees can recharge themselves & \cite{bluman2017marsbee, marsbee}\\
\hline
Kilobot & It is a low-cost swarm of small robots designed to study collective swarm behavior & Each kilobot has a programmable controller, is capable for locomotion, local communication and can sense its environment & \cite{rubenstein2012kilobot, valentini2018kilogrid, jansson2015kilombo, lopes2014application, rubenstein2010kilobot12}\\
\hline
Kobot & A circular shaped, cheap, small, and expendable robot. These features make it very suitable for various swarm robotic applications & Has IR-based short-range sensors, supports wireless and parallel robot programming and has a battery that can last up to 10 hours. & \cite{turgut2007kobot}\\
\hline
\end{tabular}
 \vspace{0.3cm}
\caption{An overview of some swarm robotic projects.}
\vspace{-0.7cm}
\label{projectscomp}
\end{table*}

\subsection{Simulators}
\label{simu}
\subsubsection*{Swarm-Sim}
\begin{figure}[ht!]
	\centering
	\includegraphics[width=0.4\textwidth]{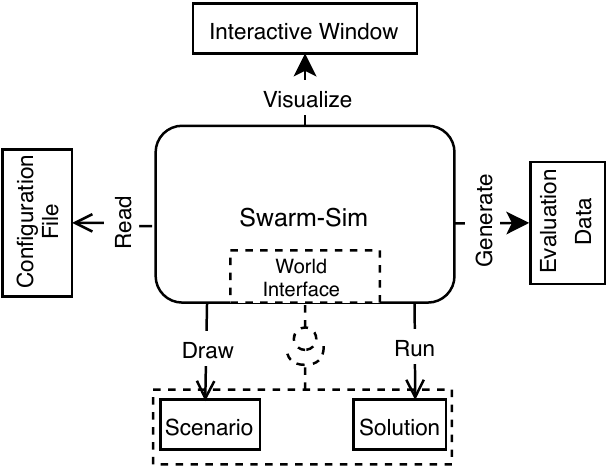}
	\caption{Swarm-sim architecture.}
	\label{archi}
\end{figure}

Swarm-Sim is developed by the technology of social networks lab Heinrich-Heine-University Düsseldorf \cite{swarmSim}, for modelling swarm robotic systems. It is programmed with Python. It is open-source and can be downloaded from \cite{swarm}. The cooperation and interactions between the robots and their environment are both modeled by the simulator. The simulator describes the modelling of operations that an agent performs in its environment and the connections between all the components involved in the same system. It provides a simple and easy configurable interface where users can create their environments according to their own requirements. The simulator is hence a simple yet flexible platform for testing and evaluating any agents in different types of environments. \\
The simulator consists of a \textit{configuration file, a scenario module, a solution module, a visualizing module} and a \textit{evaluation data}. Figure \ref{archi} shows the architecture of Swarm-Sim. Simulations in Swarm-Sim can be configured via a configuration file (\textit{config.ini}). Users can change different parameters, e.g. the maximum rounds, 2D or 3D, and etc., outside the actual code via this file. The environment is a virtual 2D or 3D space where robots and other matters are placed, and tasks are performed. The matters that can be placed into an environment include, \textit{Agents, Items} and \textit{Locations}. Agents are the robots that can move themselves, Items are objects that cannot move but can be transported by the Agents. Lastly, Locations are the coordinates in an environment where the Items and Agents can be placed. Items can act as obstacles or objects, that can be transported by the Agents. All three matters, Agents, Items, and Locations have a memory where information can be written or read. Along with movement and transporting Agents are furthermore capable of scanning their environment, creating and deleting Items or Locations, and writing and reading from the memory of other matters. 
The environments constructed by the users are stored in the \textit{Scenario} module. The solutions that are implemented are saved under the \textit{Solution} module. Swarm-sim has an evaluation data generator (\textit{csv\_ generator.py}) to produce csv files that include the results of the simulations. This file collects all the statistical data of Agents that is generated during the simulations. For example, the number of Items created or deleted by a Agent, number of Items or Agents read or written by a Agent. Along with that, the different parameters used in the simulations are also collected. For example, maximum number of rounds needed, and number of movements needed for a specific task. The collection of data is done at the end of each simulation. This makes it possible to evaluate and compare results of each simulation. The results are shown in a separated file, namely under \textit{outputs}. To see every simulation process as an animation, swarm-sim has a visualizing module known as \textit{vis.py}. This module is responsible for presenting the simulations and taking screenshots. It displays the simulations graphically in a separate window. This gives users the opportunity to see an animation of their implemented scenarios and solutions and examine the Agent behaviors when accomplishing different tasks. There is some research published with the simulator Swarm-Sim. Such as coating an object with help of swarm \cite{coating, leader}, the flocking movement of swarms in a triangular graph \cite{flocking}, a simulation of marking any given field with obstacle, and three more \cite{phototactic, antmill, oppnetswarm}.

\subsubsection*{SCRIMMAGE}
Simulating Collaborative Robots in Massive Multi-Agent Game Execution (SCRIMMAGE) \cite{demarco2019simulating} is a 3-dimensional, open-source simulation environment for the testing and comparing of mobile robotics algorithms and behaviors. SCRIMMAGE takes its inspiration from the properties that the stage project \cite{vaughan2008massively} offers, namely the simulation of multiple robots, the plugin interface for robotic control, models for locomotion, sensors, and the autonomous communication amongst robots. The main aim to develop such a simulator was to overcome the lack of open source 3-dimensional simulators, that can simulate not only normal walking robots, but also aerial robots. SCRIMMAGE has a batch processing, and it can analyze simulation results. Multiple plugins are provided by SCRIMMAGE, including autonomy, motion model, controller, sensor, entity interaction and metrics plugins. Visit the SCRIMMAGE home page \cite{Scrimmageweb} for details on these plugins. These plugin interfaces provide different levels of simulation fidelity. Additional plugins can also be written, for example, for collision detection. The simulations can be paused and stopped by the research. This feature decreases the time for the development of autonomy. SCRIMMAGE not only simulates aerial robots but also ground and water robots. 

\subsubsection*{Robot Virtual Worlds}
Robot Virtual Worlds \cite{RVW} is a simulator for modelling virtual robots. This is a unique and useful way for new research and students to learn and test programming without any physical hardware. It supports the programming language ROBOTC, that is used to implement NXT-G, EV3, LABVIEW robots. Normally, research and developers, first develop physical robotic hardware and then simulate the code. So, the modelling and simulations are done after the development of real robots. With robot virtual worlds, the modelling and testing is done without any requirement of real robots. Codes can be simulated, and functions can be tested. So, overall implementations can be optimized via this simulator. Due to this reason, robot virtual worlds is very practical for students who often do not have many resources to build robots for testing.

Robot virtual worlds simulator offers many useful tools that help students to design anything according to their own requirements. These tools include, level importer, model importer and the measurement tool kit. With the help of the level importer, one can place many all-day use items into tables, such as chairs, tables, balls etc.. With such placing they can build different environments for testing and put various challenges for their objects. Objects can also be imported and created according to students' requirements, in a three-dimensional space via the model importer. The most useful tool is the measurement tool that allows to detect the path planning of objects. This way students do not need to guess the paths, that the robots are going to choose or how far they are going to go. 

\subsubsection*{Webots}
Webots \cite{olivier2004cyberbotics} is a widely used, high-quality professional mobile robot simulator. It is mostly used for educational purposes. It has a development environment that can be prototyped to permit the users to create environments in the 3-dimensional space and with features like, volume, mass, links, friction coefficients and so on. Users can create and modify the environment by adding or removing robots. These robots can have different forms and sizes such as robots with legs or wings. Furthermore, a user can add sensors and actuators to the objects, such as light sensors, wind sensors, cameras etc.. \\
The simulation consists of two main components, namely the environmental and the multiple controller implementations. The environment is basically a \textit{world} that consists of robots and other items that are included by the users. In addition to the normal robot controllers, there are some \textit{supervisor controllers} too. Supervisor controllers are robots with extra rights and can carry out activities that are not possible for normal robots. For example, moving a robot from one place to another, capturing a screen shot or a video of the simulation and so on. These robots are affiliated with the normal controllers and can also be implemented in the same languages in which the controllers are. Furthermore, physical plugin can be used, that can help to change physical behavior of the simulation. Other features of Webots include that Webots physical engine is implemented with the open source 3D dynamics physics library Open Dynamic Engines \cite{ODE}, and for implementing the robots, various programming languages can be used, like C, C++, Java and some others via TCP/IP. Robot controllers can also be implemented in several languages, such as C, C++, Java, MATLAB and Python. The simulation process is fast due to the reason that there are no standards being published for the evaluation of performance or quality. To help users, Webot comes with a complete documentation about how to use and work on it. Furthermore, it has also various examples for control programs and robot models.

\begin{table*}[t]
	\center
	\begin{tabular}{|P{1.5 cm}|P{4.0 cm}|P{2.0 cm}|P{0.6 cm}|P{2.0 cm}|P{1.7 cm}|P{0.5 cm}|P{0.7 cm}|P{1.2 cm}|}
		\hline\textbf{Simulator}&\textbf{Objective}&\textbf{Developers}&\textbf{Open source}&\textbf{supported languages}&\textbf{Supported OS}&\textbf{2D or 3D}&\textbf{Status}&\textbf{Researches}\\
		\hline
		\hline
		Swarm-Sim & A round based simulator developed for modeling swarm robotic systems in a 2D/3D environment. & Heinrich Heine University of D{\"u}sseldorf \cite{swarmSim} & Yes & Python & Linux, MacOS, and Windows & 2D and 3D & Active& \cite{coating, leader, flocking, marking, phototactic,   antmill, oppnetswarm}\\
		\hline
		Player and stage & offers free software for robots, sensors and actuators research & Brian Gerkey \cite{briangerkey}, Richard Vaughan, Andrew Howard, and Nathan Koenig & Yes & any language & Linux, Solaris, BSD and MacOSX & 2D & Last update 2010 &\cite{vaughan2007reusable,gerkey2001most, gerkey2003player,vaughan2008massively}\\
		\hline
		Gazebo & offers opportunity to simulate robotic swarms accurately and efficiently in various indoor and outdoor environments & Open Source Robotics Foundation (OSRF) \cite{OPR} & Yes & mostly ROS (Robot Operating System)\cite{Ros} & MacOS, Linux and Windows (a binary package available only for Linux) & 3D & Active &\cite{meyer2012comprehensive,koenig2004design}\\
		\hline
		Robot Virtual Worlds & High-end simulation environment for students to learn programming & Robomatter Incorporated \cite{Robomater} & No & ROBOTC \cite{Robotc} & Windows and Mac & 3D & Active &\cite{RVW}\\
		\hline
		Teambots & Offers java classes and APIs to support research in mobile multi-agent systems & Georgia Tech’s Mobile Robot Laboratory \cite{Georgiatechmobile} & Yes & Java & Windows, NT, Solaris, SunOS, MacOS, OS X, Linux and IRIX & 2D & Last update 2000&\cite{balch2001teambots,balch2002teambots,balch1998behavioral}\\
		\hline
		V-REP & A universal simulator with integrated development environment, where each item can be controlled individually & Coppelia Robotics \cite{vrepcoppel} & Yes & C, Python, C++, Java, Lua, Matlab, Octave and Urbi & MacOSX, windows and linux, & 3D & Active&\cite{freese2010virtual,olivares2015vision,peralta2016development,rohmer2013v}\\
		\hline
		ARGoS & Aims to simulate heterogeneous robotic swarms in real-time & Developed within the swarmanoid project \cite{pinciroli2007swarmanoid} & Yes & ASEBA scripting language (others are under study) \cite{magnenat2011aseba} & Linux and MacOSX & 2D and 3D & Active&\cite{allwright2018argos,pinciroli2011argos}\\
		\hline
		Webots & high-quality professional mobile robot simulator  used for educational purposes & Cyberbotics Ltd. \cite{WEBOTSOPENSOURCE} & Yes & C, C++, Java and from third party software (via TCP/IP) & Windows, Linux and MacOSX & 3D & Active&\cite{michel1998webots,olivier2004cyberbotics,wang2000developing}\\
		\hline
		Workspace & An Offline Simulation and programming platform. Offers simulation solutions for industrial and educational purposes & Watson Automation Technical Solutions Ltd. \cite{Watsolutions} & No & Many robotic languages e.g. AB G-Code and Adept V-Plus & Windows & 3D & Active&\cite{WORK}\\
		\hline
		OpenHRP & A virtual platform to investigate humanoid robotics & AIST \cite{AIST} & Yes & C, Python, C++, Java & Linux, Windows & 3D & Last update 2012&\cite{kanehiro2004openhrp,mittal2015implementation,hirukawa2004humanoid}\\
		\hline
		SCRIMMAGE & Used for the testing and comparing of mobile robotic algorithms and behaviors & Georgia Institute of Technology \ & Yes & Python, C++ & Linux, MacOS & 3D & Active&\cite{demarco2019simulating} \\
		\hline
	\end{tabular}
 \vspace{0.3cm}
\caption{Comparison and differences between several swarm robotic simulators.}
\vspace{-0.7cm}
\label{similatorcom}
\end{table*}

\subsection{Real life Applications}
\label{appreal}
\subsubsection*{Agriculture}
\textit{SAGA}\\
\textit{SAGA} \cite{trianni2016saga, albani2017monitoring, SAGA} is an experimental platform for agricultural swarm robotic systems. The focus of the project is to demonstrate the importance of swarm robotics in the field of precision farming \cite{blackmore1994precision}. The experiments show a swarm of \textit{unmanned aerial vehicles} (UAVs) deployed in a field with the aim to monitor the field and perform weeding. UAVs are aerial drones developed by avular.B.V \cite{AVULAR, PRECISIONSCOUT}. These UAVs are enhanced by the SAGA platform by adding some extra features and components. The components and features include on-board cameras, vision processing, radio communication systems and protocols that will offer support for safe swarm operations. With the help of the on-board vision system, weed detection will be done on-board to count the number of weeds in an area and detect areas where weed is in sufficient amount. The UAVs first search for volunteer potatoes and sugar beets so that harmful areas can be detected. After that, the process of weeding is started. The UAVs fly very close to the field to get images and detect weed in the plants. The project focuses on only considering areas where there is enough weed. Multiple UAVs search for weed and in case other tasks are required, like micro-spraying on some plants, some UAVs can deal with these extra tasks. This also shows the importance of having multiple UAVs rather than just a single UAV. \\
Experiments were conducted with the SAGA platform with different parameters to see the influence of these parameters on the main task. These parameters include the number of UAVs deployed, weed detection rate and the distribution of weed into the fields. These parameters were tested in randomly generated fields. The results showed that the time to detect weed depends highly on the weed detection rate. The system also dealt well with low weed detection rates and showed good scalability. \nocite{SWARMFRARM}

\subsubsection*{Industrial}
\textit{FIBERBOTS}

\textit{FIBERBOTS} \cite{kayser2018fiberbots} is a very recent project (developed in 2018) where robots act as swarm fabricators and can design fabricated structures. They work in parallel to form tubular forms. Robots pull fiber and resin from a ground-based storage (fiberglass spool) and wrap it around their own bodies. They build the structures by winding the fiber into tubes around themselves. The building of the structure consists of several steps. A winding arm pulls the fiber and resin from the ground-based storage. These both materials are mixed in the nasal and then winded itself. In the next step, the fiber is hardened with the help of ultraviolet light. After hardening the fiber, the fiberbot inches itself on top of the hardened fiber, with the help of its tiny motor and wheels. This process repeats itself until the aimed structure is built. The winding patterns can be changed to make the structure look different. The robots do not run into each other, because they communicate with each other via a computer network and know each other's current states. There is no need to command the robots, instead the system uses a flocking-based design protocol \cite{reynolds1987flocks} for the structure formation for informing the robot trajectories. \\
A very recent hotel project in China named as \textit{FLYZOO FUTURE HOTEL} \cite{alibabaflyzo}, is known as the most modern hotel in world. The hotel works with robotic hotel staff that are responsible for almost every task in the hotel, from check in to food delivery and all the other stuff that is needed in a hotel. Hotel guests check in via scanning their passports via robot, that also has face recognition to provide the guests access to all other important stuff, like using the lift and opening the rooms. Every room has its own virtual assistant that can be ordered to for example, close the windows, closing the curtains, playing music, or ordering food. The bar has its own robots that make drinks. Guests can choose drinks via an app, and they are then prepared and delivered by the robots. However, human staff is still needed for some tasks, like the preparing of food. It is a new concept that is in its evolving stages. Maybe in the near future we will see hotels only with robots and no human staff.

\nocite{werfel2014designing, TERMES4, AMAZON, DOMCHINA, OCADOWA}
\subsubsection*{Medical}
\begin{figure}[ht]
	\centering
	\begin{minipage}[t]{0.45\linewidth}
	\centering
  \includegraphics[width=1.0\textwidth]{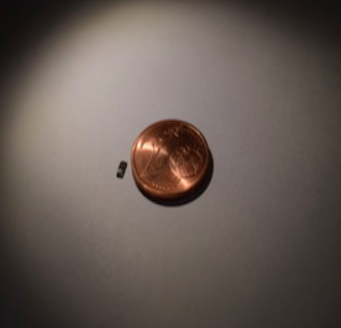}
	\caption{The millirobot as compared to the size of a coin \cite{MILLIROBOT}.}
	\label{millirobot}
	\end{minipage}
	 \hfill
	 \begin{minipage}[t]{0.45\linewidth}
	\centering
  \includegraphics[width=1.0\textwidth]{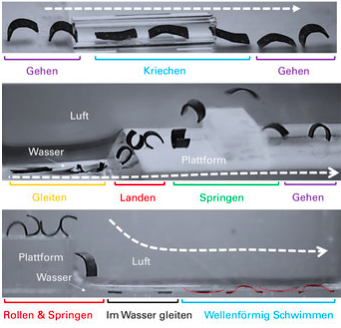}
	\caption{The type of movements the millirobot can perform \cite{MILLIROBOT}.}
	\label{millirobot1}
	\end{minipage}
\end{figure}

A very attractive field of research in swarm robotic systems is cancer treatment. Even though the technology is getting advanced and the treatment of almost all types of cancer is available. But the side effects that these treatments can cause are still a huge problem. The main problem is the attacking of healthy cells. This problem can only be solved with a treatment that only considers unhealthy cells. Swarming nanorobots is one such treatment system that can be used to kill cancer cells while not causing any damage to the healthy cells. The nanorobots can be designed in such a way that they only target the cancer cells. They should be able to navigate in their environment, that is the human body and search for and recognize specific items, that are the cancer cells. The treatment can be of different types. The nanorobots can either search for cancer cells and destroy them by injecting the medicines, or they can drill into the cells \cite{garcia2017molecular} and bust them without the need of any drug. When a swarm of such nanorobots move in a cancerous body, they can kill the cancer cells very fast. \\
Scientists from the Max Plank Institute have developed millirobots \cite{MILLIROBOT, MILLIROBOTMAX}, that have various features. They can walk, crawl, roll, swim, and transport items. This means the robots can move on land as well as water (see figure \ref{millirobot1}). The length of this millirobot is just four millimeters (see figure \ref{millirobot}). The different kinds of movements are made possible with the help of magnetic micro particles embedded in their silicon rubber body. The robots can be controlled by the external magnetic field. Researchers can change the form of the robot by changing the strength and direction of the magnetic fields. This way, the robots perform spontaneous actions like humans. For example, jumping across any items in their way, crawl through narrow areas and so on. The research aim to make these millirobots able to transport drugs in various environments, including the human body. 
\nocite{hougen2006swarm, cavalcanti2008nanorobot, cavalcanti2007nanorobot, cavalcanti2005nanorobotics, al2011swarming}
\subsubsection*{Hazardous Zones}
\begin{figure}[ht!]
	\centering
  \includegraphics[width=0.4\textwidth]{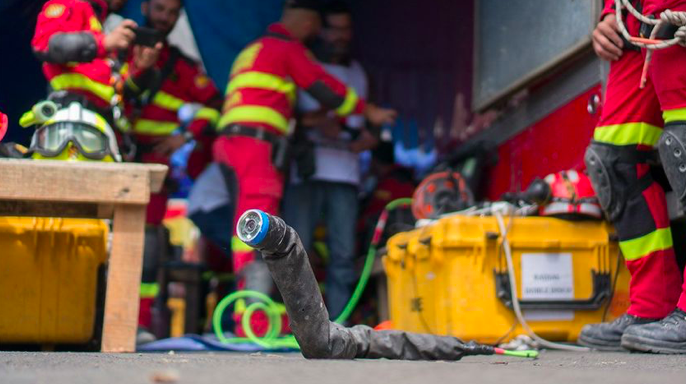}
	\caption{The snake robot that took part in the searching of survivors after the 2017 Mexico earthquake \cite{SNAKAEFOT}. }
	\label{snake}
\end{figure}

The Howie Choset’s research group at Carnegie Mellon University in Pittsburgh, Pennsylvania created snake robots \cite{SNAKEBOT} that can pass through tight tunnels and paths where humans or other machinery cannot pass through. These robots are made up of metal and can crawl, swim, climb and perform many other behaviours. They are around five centimeters thick and about one meter long. With the help of 16 joints, they can pass through tight areas very easily. The snake robots are also equipped with light and cameras so that the footage of every area is clear and can be seen at the control center. These robots helped in the searching of survivors after an earthquake hit Mexico in 2017 (see figure \ref{snake}). However, there were no humans found during the operation by the snake robots. The researchers aim to enhance the system and design of their robots. For example, by adding various sensors and new visualisation tools. The research group is working on several other robots, including trunk snake \cite{TRSNAKEBOT}, medical snake \cite{MEDSNAKEBOT}, fullabot \cite{FLSNAKEBOT} and others. 

\nocite{wang2017locating}
\subsubsection*{Military}
Endeavor Robotics \cite{ENDEAVOR} developed several robots that are being used by the US military for various tasks. The cobra robot \cite{KOBRA} is used for lifting heavy objects. It can lift to 150 kg of weights. The packbot \cite{PACKBOTF} is used for bomb disposals. Boston Dynamics designed AlphaDogs \cite{BOSTONSDY}, that are robots that look like dogs and are used to transport heavy loads for soldiers. It can carry up to 180 kg of weight and can walk up to 20-miles. The interesting part is that these robots do not need to be controlled to tell them in which direction they need to do. With the help of computer vision, they automatically follow their leader. 
\nocite{young2016survey, krishnan2016killer}
\subsubsection*{Household}
Adrian Perez Zapata from Colombia presented a design "MAB" for an automated cleaning system \cite{ELECTlux} that consists of a huge number of mini flying robots. These robots are equipped with cleaning essentials, and they fly around in homes to clean most of the surfaces. They can also carry dust and dirt particles from the surface and depose them. Environmental information is exchanged among the mini robots, via direct communication. This system concept won the first position in the Electrolux Design Lab competition in 2013. 

\nocite{IJIR457, prabakaran2018floor, pandey2014technological, fiorini2000cleaning, fiorini2000cleaning}

\subsubsection*{Astronomy}
One of the interesting new field for swarm robotic systems is Astronomy. Research are trying to find ways how robots can help them explore the non-earth bodies like Mars, sun and the moon. The concept of \textit{"robotic telescope"} was introduced in early 1980s \cite{trueblood1985microcomputer}. Back then they were costly and due to resource limitations, difficult to develop. Research in Liverpool John Mores University developed the eSTAR project \cite{allan2004estar}, that is a robotic technology to carry or telescopic tasks. It is a multi-agent system that is made up of a group of heterogeneous robotic telescopes. Another project that deals with heterogeneous robotic telescopes is \cite{mason1994collaborative}. The agents used are autonomous and can coordinated with each other. The authors developed protocols for the coordination and collaboration among the agents. They also established a programming language and an agent architecture for these protocols. An overall view of using robots in the field of astronomy is given by \cite{baruch1992robots}. The author discusses the development of robotics in industrial and astronomy area, the reasons to use robotics, the advantages, and the shortcomings. All the aspects are discussed in detail. 

\section{The future}
\label{fut}
\begin{figure}[ht]
	\centering
  \includegraphics[width=0.4\textwidth]{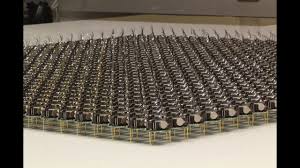}
	\caption{A swarm of a thousand robots \cite{swarmthou}.}
	\label{swarm}
\end{figure}

The field of swarm robotics has developed very strongly in the past years. From simulation platforms to real life applications, all aspects are evolving day by day. Many systems are already in deployment (some of them discussed in section \ref{appreal}) whereas many are under construction and in testing phase. However, the development of swarm robotics in real life applications is still in its infant stages. Most of the real-life systems are used for tasks that do not necessarily have a human contact because the systems need to be very secure before deploying them in areas with direct human contact. Most tests for swarm robotic systems is limited to simulators. These simulators are mostly not able to take real life circumstances into account. Due to which, a system that has been tested successfully in a simulator is not necessarily also successful in the real life. Also, the projects that test these systems with real robots are limited due to many aspects, like the lack of hardware or software components and most importantly the cost factor. \\
The modern research and deployment of several swarm robotic systems took off in the early 2000s. SWARMBOT (see section \ref{proj}) being one of the early projects in this field, that aimed to build fully autonomous robots namely, the \textit{s-bots}. These \textit{s-bots} could grasp objects and transporting them via assembling with other s-bots. In 2010 the kilobot \cite{rubenstein2010kilobot} project was developed, that consisted of thousands of small and simple autonomous robots (see figure \ref{swarm}) capable of various tasks via aggregation and self organisation amongst robots. TERMES \cite{werfel2014designing,TERMES4} and FIBERBOTS (see section \ref{appreal}) are two very recent projects under the industrial applications of swarm robotics. Both projects are aimed for development purposes. Industrial automation is growing very vast. Many countries including Japan, USA, Germany, and China are developing more and more application to automate industrial applications. According to the International Federation of Robotics (IFR) more the 1.4 million industrial robots will be installed in factories by 2019 \cite{WorldRoboticsReport}.

These research show that the evolution of swarm robotics is not only growing but growing in the direction of huge similarity with natural swarms. The idea of swarm robotics takes its inspiration from the natural swarms. These natural swarms are autonomous, have no direct communication, self-organized, coordinated, and cooperative. These are some of those features that make natural swarms build complex swarm behaviours that result into the accomplishment of several complex tasks. So, the focus of building swarm robotic systems remains of this fact, that the systems build should be able to gain complex goals while staying simple. However, the building of systems with such properties is not an easy task. This leads us to the difficulties and challenges that are still not solved or not solved completely. Solving these issues should be the focus in the future, in order to develop systems that have almost exactly the same features that natural swarms have and are capable of doing all the tasks that we humans require them to do. This could be very helpful to get a step closer to the development of more swarm robotic applications in real life. The issues that we discuss are categorized into two classes namely, the \textit{hardware} and \textit{software} issues. The next two sections discuss these two development issues. Section \ref{possfut} gives an overview of some possible applications, according to my knowledge, that could be accomplished in the future. 

\subsection{Hardware and Software Issues}
The hardware and software of swarm robots must be constructed in such a way that it supports the three essential features, flexibility, scalability, and robustness, of a swarm robot. Additionally, an important aspect is the cost factor. Flexibility should be achieved by the self organising as well as distributive behavior of robots. This means they should be able to aggregate and distribute as required by the given tasks. The systems must be able to work with different sizes of groups. This leads to the scalability of the system. Robustness is how the robots cope with the changing number of members in their swarm. That means they should still work with the same efficiency even if some members are failed to operate. The failures should also not have any effect on the performance of the whole system. We suggest following hardware requirements for swarm robots: 
\begin{itemize}
\item{Add-On design, so any hardware modification can be made easy and quick.}
\item{Equipped with a minimum sensors that are able to sense things like items, substances, or recognizing their own kind.}
\item{Mobility components e.g. wheels to move from A to B}
\item{Grasping components to grasp items or its own kind.}
\item{Wireless communication components, e.g. WIFI, to communicate with others}
\end{itemize}
These are some of the hardware requirements. There are still many other open issues that need to be solved before we can build a swarm system that actually works like a natural swarm. Many systems offer several features discussed above. For example, kilobot \cite{rubenstein2010kilobot} offers the feature of being simple, small, autonomous, and cost effective. SWARMBOT \cite{dorigo2005swarm} built s-bots that are self-assembling and self-organizing robots, the I-SWARM \cite{woern2006swarm} project built tiny robots (3*3*3 mm) and aimed to create the largest robotic swarm with these mini robots. Some systems tested the aspect of heterogeneity in swarm robotic systems. Like the swarmanoid project that is the successor project to SWARM-BOT and has extended the work of SWARM-BOT by working on a three-dimensional environment. The focus was to build heterogeneous swarms. The UB-swarm project \cite{articleubswarm} also built a swarm of heterogeneous robots.\\
Thus, in our opinion we suggest some software requirements include the implementation of:
\begin{itemize}
\item{An operating system}
\item{Communication Protocols}
\item{Swarm Algorithm that is capable to be flexible, robust, and scalable}
\item{Other features like, navigation, scanning, obstacle avoidance and so on.}
\end{itemize}
A very crucial task is to build robots that can operate in natural environments. Because in contrast to simulators and test environments, natural environments are not fully known, and the changes are spontaneous. So, robots must be able to cope with these spontaneous changes and work effectively in unknown environments. Even more difficult it is when mini robots are aimed to work in places where humans have no access to, for example, collapsed buildings, radioactive places or inside a human body. Nanorobots are aimed to work inside human bodies. They should be able to work, explore and navigate in the human body in the most effective way, so that the required task is accomplished successfully, without any damage to the human body. \\
The communication also plays a very important part while building swarm robotic systems. In natural swarms the communication is explicit, that means not direct. Natural swarms communicate mostly via pheromones, that are chemical substances to signal other members of a swarm. Providing such type of communication for swarm robotics needs complex algorithms. \\
The developed systems lack many features. So, there is no such system, up to my knowledge, that offers all the important features (stated in section \ref{properties}) that are listed as the 'essential' features for a swarm robotics system. The future goal should be to take the hardware and software issues, listed above, into account and build such systems that look and act in the most similar way as natural swarms. This will be a step forward towards building effective real-life swarm robotic systems that are simple, affordable, and yet efficient enough.\\
Researchers have dreamed since a long time to build swarm robotic systems that are fully autonomous and capable of all these essential features. Many projects show that these requirements are being accomplished more and more (some projects listed in \ref{appreal} but as the technologies grows the field is getting even more complex. The environments are getting complex, the tasks are getting more difficult, so are the requirements of building effective swarm robotic systems for real life. Robots need to accomplish more complex tasks with more difficulties but with even better efficiency. The current development of swarm robotics shows that there is lack of performance by swarm robotic systems deployed in real life scenarios, due to the limitations in the software or hardware of the robots. Hence, there is a lot to be accomplished in the development of swarm robotics software and hardware. 

\subsection{Possible Future Applications}
\label{possfut}
\subsubsection{Nanorobots in Medicine}
The nanotechnology is becoming one of the most promising applications of swarm robotics. They are being used for many different reasons, medicine being the most important one. Scientists are finding ways to develop nanorobots technologies that would be able to kill many diseases. Killing cancer via these nanorobots is one of the main tasks that scientists are aiming to achieve. Nanorobots should be able to navigate inside a human body and look for the diseased cells. The killing can be done directly by drilling into the cells \cite{garcia2017molecular} or through a medicine that the robots carry with them. The Max Plank Institute has built nanorobots \cite{MILLIROBOT, MILLIROBOTMAX} that they aim to use for curing cancer. They have also developed nanorobots, very recently in 2018, that are aimed to propel through the eye \cite{nanopropel}. Until now, such nano propellers were only able to travel through fluids or models but not through tissues. But these newly developed nanorobots can travel through real tissues and can deliver the required medicine exactly where it is needed in the human body. \\
These ideas and inventions are a groundbreaking revelation in the field of medicine. But so far there have been no human trials. The nanotechnology can be used for many other diseases that do not have any cure yet or not an effective one for example, HIV. There is no specific medicine that can cure HIV. Patients are given drugs that can increase their life span but not erase the virus completely. Nanorobots can be used to destroy HIV particles. Nanorobots can be designed with such capabilities that they are able to find these HIV particles, that are around 60 times smaller than the red blood cells and destroy them. However, there are several aspects that need to consider. Designing such nanorobots can be very costly and complicated. They also need to be very precise and accurate so that they only target the diseased cells and not the healthy one and hence not causing any harm to the human body.
\subsubsection{Military and Navy}
The US military is already using swarm robotics systems for a lot of tasks. Some of these tasks and projects are mentioned in section \ref{appreal}. However, these robots are mostly used for tasks that are not possible for humans or are very difficult to accomplish. For example, the transportation of heavy products, disposal of chemicals or searching into places where humans may have no access. The next step can be the use of robot swarms as armed forces. That means the replacement of human army with army of swarm robots. The development of such robots has already begun. DARPA is already trying to develop weaponized swarms of robots that can work in replacement of human armed forces \cite{darpadrones,darpamilitary}. \\
Not only the army but such swarms can also be used by the navy. For example, underwater data can be collected via robots. These robots should be autonomous and able to interact with other members of the swarm and hence achieving tasks via their swarm intelligence and swarm behavior. Tasks can be for example, monitoring, searching for specific items or just imaging the underwater environment. A crucial part in this research would be to find an effective way of communication amongst the robots in the underwater environment. Another issue would be the cost factor. Building such complex systems would need many other features than just scalability, flexibility, and robustness. 
\subsubsection{Replacing of Labour}
Swarm robotics can be used to replace labor. That means, instead of having human labor in for example stores, hotels or industries, robotic swarms can be used. Some swarm robotics industrial projects are already in use \cite{OCADOWA,DOMCHINA,AMAZON}, where robots are responsible for packing groceries, transporting or sorting items. However, it is important that these robots have Artificial Intelligence, only then would it be possible to replace humans with automation. Many other jobs can be replaced completely, or human assistance can be decreased with the help of robots. The jobs do not need to be complex but also simple jobs can be replaced by robots and hence decreasing labor costs. For example, robotic swarms can be hired as assistance for cleaning in a factory or store, for helping customers in transporting heavy items to their cars, helping in making food or drinks, or in restaurant for delivering food items to tables. Such simple tasks are possible to attain by swarms of simply designed and cost-effective robots. 

\subsubsection{Snake Robots for many Purposes}

Swarm robotics can play an efficient part in surveillance systems. Areas that are unknown, too dangerous, or not reachable for humans can be monitored via swarm robots. Single robots are not an efficient choice for such tasks, due to many reasons, for example, the area is too big to be covered by a single robot. Also, stationary robots are not an ideal choice, the robots need to be mobile to get a good image of the environment. Swarms of snake robots can be a very effective choice for this task. Snake robots can be used for a vast range of tasks, like in medicine, industry, manufacturing, and others. The task of surveillance being one of the most important from them. Snake robots are self-contained robots, that can access places that are not accessible for humans. For example, looking for survivors in a collapsed building or in a radioactive environment. Due to their slender bodies and many joints that they have, they are very flexible and can pass through very small places and can even swim. Such robots can be used for rescue missions and increase the success chances. They can be used as spy cameras, for conventional open surgery, surveillance under water, climbing mountains or buildings and many more tasks. 

\subsubsection{Robots in the Space}
Our solar system has been studied since a hundred to thousands of years. Humans have already traveled to Moon but there is yet a lot to explore about it and about the rest of our solar system that consist of many other planets, asteroids, and stars. The exploration of Mars has been a mission for many years now. NASA aims to send humans to explore the planet since very long but there is yet no project that states when it is going to happen. According to Chris Hadfield, a former International Space Station commander, NASA could have sent people to Mars decades ago, just like it sent people to the Moon. But the reason it did not happen is that the chances of death on Mars were extremely high \cite{nasaschocker}. This can lead us to plan projects that include only robots performing the tasks of space or planet exploration. In the manned missions on moon, most space exploration tasks are already done by robots. We can exclude the sending of astronauts and make the missions completely performed by swarms of robots. This would not only be a cheaper task but also non-risky for human lives. There have been a lot of deaths and injuries of astronauts in space, during training, launch phases and atmospheric re-entry. NASA had its first disaster in 1967, known as the Apollo 1 disaster, where the whole crew lost their lives due to a fire in the cockpit. \\
The research and development of swarm robotics can be a groundbreaking revelation for space missions. Planets can be explored via hundreds to thousands of robotic swarms, that are designed simply and with low costs. These robots should be designed in such a way that they can cooperate and are able to sense different environmental gases on these planets, look for water and food resources, take images and videos, collect samples, and send this information back to earth. 

\subsubsection{Inspection}
Swarm robots can be used for inspection purpose in the industrial field. Engineered industrial structures, small to large, like pipes, turbines, tanks, boilers, and vessels need regular inspections. These inspections can be dirty and dangerous. Inspecting successfully is also not always the case because humans may not be able to access all the parts. Here, swarm robotics can play an important part. The resources and costs both will be decreased, and the danger factor would also be excluded by taking the humans out. 

\section{Related Surveys}
\label{relsur}
The research work in the direction of multi-robotics and swarm robotics is driven by different inspirations and have hence difference goals and objectives. Some of the research work deals with the overall idea of multi-robotic systems, their requirements and so on \cite{balch2002robot, cao1997cooperative}, whereas others deal with specific multi-robotic projects, swarm behaviors and platforms \cite{mataric1997reinforcement, mataric2001learning}. \cite{balch2002robot} deals with the aspect of heterogeneity amongst robots in multi-robotic systems. A detailed review of mobile robots is provided in \cite{cao1997cooperative}. In \cite{mataric1997reinforcement} the authors elaborate the idea of reinforcement learning by collective robots in multi-robotic systems. Another research paper \cite{mataric2001learning} by the same author, extended the previous work and increased the number of robots in the experiments and discussed other types of swarm behaviors. Some research work on specific swarm robotic projects and platforms include SWARBOTS \cite{dorigo2005swarm}, I-SWARM \cite{seyfried2004swarm}, Webots \cite{olivier2004cyberbotics} and kilobot \cite{rubenstein2012kilobot}. Some projects and simulators are discussed in chapter \ref{proj} and \ref{simu}. \cite{hayes2002many} used the Webots simulator to simulate their experimental tasks. \cite{martinoli1999swarm} provided an extensive research work on autonomous robots. The work was conducted using the Khepera robotics platform \cite{Kteamkhepera}.\\
Nanorobots, that are a popular topic of interest nowadays, was discussed by \cite{lewis1992behavioral}, in 1992. The authors dealt with the idea of using \textit{pheromones} in order to organize nanorobots. Pheromones are chemical substances that are used by animals or insects, such as ants, to signal other members of its species. The aim was to destroy a common goal via some specific rules. The robotic behaviors were simulated and results were positive, that means the robots were able to destroy their target via signaling each other.\\
There is a huge list of surveys available for multi-robotic systems \cite{yan2013survey, choset2001coverage, mohamed2008middleware, iocchi2000reactivity}, multi-agent systems \cite{stone2000multiagent, 10.1007/3-540-58855-8_1, 10.1007/3-540-49057-4_21}, as well as surveys that specifically deal with the topic of swarm robotic systems. Under the category of swarm robotics, there are surveys that deal with every aspect of swarm robotics, like features, requirements, projects and many more categories and some deal with specific aspects like security challenges for swarm robotic systems \cite{4976621abc}, building a human-swarm system \cite{7299280bcd}, survey on the research directions in swarm robotics \cite{inproceedingsextensivere, articlesurveystudies}, possible applications for swarm robotics \cite{csahin2004swarm}, swarm robotic algorithms \cite{SENANAYAKE2016422, articlealgoswarm, Shlyakhov_2017} and many other aspects \cite{articlealgodistr, brambilla2013swarm, articleinteractive}. Some surveys that give a global view on the field of swarm robotics, by discussing almost all important aspects, like classification of swarm robotics, features, rules, requirements and others, include \cite{10.1007/978-3-642-30976-2_68, inbookyhingtan, articleintroduction, tan2013research}. Most of the surveys, deal with the aspect of swarm robotics in a way of giving an overall picture for this field. This means, they discuss the important features, inspirations, goals etc., but there is a little about the history of this field and the connection of the historical aspects with the present and future aspects. The main focus of this paper was to give an overview of swarm robotics starting from the very beginning to the future perspectives. 

\vspace{0.150cm}
\section{Conclusion}
Swarm robotics is the study of designing groups of simple and autonomous robots that can cooperate and coordinate with each other. They can self-organize and operate without a central entity. The designing of such groups of robots is not an easy task. The costing factor also plays a very important part in the designing. Researchers and scientists aim to build simple and cost-effective robots that can accomplish complex tasks via their collective behavior and intelligence, namely the swarm behavior and swarm intelligence, respectively. To achieve desired swarm behaviors, complicated algorithms are needed, and the factors of scalability, flexibility and robustness need to be considered. The research of swarm robotics began in the late 1900s and developmental work stared in early 2000s. This field has evolved vastly in the past few years, from simulators to real life projects, every aspect has a lot of research- and practical work to be considered. \\
This paper has discussed the field of swarm robotics, starting from its beginning to its current position. The concept of swarm robotics is explained in detail, from features to advantages, shortcomings, and classification. Most of the early swarm robotic algorithms were tested via simulation platforms and some of the tests were also conducted with real robots. Some recent and old simulation platforms and projects are discussed. Real life applications of swarm robotics are recently being developed in several fields like agriculture, household and medical. These real-life applications are enlightened in this work. This paper also discusses the future scope of swarm robotics, by giving an overview of some topics and applications that can play an important part for the evolution of swarm robotics. We can conclude from the huge amount of research work that has been done in this field, that there is still a long way to go, to develop systems that look and act like natural swarms, that is the main motivation behind the field of swarm robotics. Even though the testing of such systems has already begun in real life scenarios, there is still the need of much more examination to make sure no such point is left out of consideration, that is not only important for the effectiveness of the systems but also for the security of human beings, because such systems are eventually developed for the assistance of human beings.

\section*{Acknowledgments} 
We would like to thank the Arab-German Young Academy of Sciences and Humanities for funding this research as well as our collaboration partner Dr. Ahmed Khalil for the valuable discussions.
\vspace{-0.3cm}
\bibliographystyle{IEEEtran}
\vspace{-0.3cm}
\bibliography{swarm-survey}

\end{document}